\documentclass[11pt,letterpaper]{article}
\usepackage[T1]{fontenc}
\usepackage{times}
\usepackage{microtype}
\usepackage[letterpaper,textwidth=6.45in,top=.95in,bottom=1in]{geometry}
\usepackage{amsmath,amssymb,graphicx,booktabs,tabularx,array}
\usepackage[numbers,sort&compress]{natbib}
\usepackage{pifont,placeins,caption,xcolor}
\usepackage{tikz}
\usetikzlibrary{arrows.meta}
\definecolor{inklink}{HTML}{245E68}
\usepackage[colorlinks,allcolors=inklink,bookmarksnumbered]{hyperref}
\hypersetup{pdftitle={RigRecast: Large-Baseline Re-observation of Dynamic Driving Scenes}}
\newcommand{\method}{RigRecast}

\newcommand{\metrichead}{Method & IoU $\uparrow$ & PSNR $\uparrow$ & SSIM $\uparrow$ & LPIPS $\downarrow$ & FID $\downarrow$ & FVD $\downarrow$}

\hypersetup{pdfauthor={Yizhao Han, Xuan Yang, Kai Liu, Yifei Tong, Tianxing Shi, Ziyang Huang, Hao Zhu, Xiao-Xiao Long}}
\newcommand{\PublicByline}{%
  \par\vspace{0.55em}
  {\normalsize
  \href{https://openreview.net/profile?id=~Yizhao_Han1}{Yizhao Han}\quad
  \href{https://openreview.net/profile?id=~Xuan_Yang25}{Xuan Yang}\quad
  \href{https://openreview.net/profile?id=~kailiu6}{Kai Liu}\quad
  \href{https://openreview.net/profile?id=~Yifei_Tong2}{Yifei Tong}\\[2pt]
  \href{https://openreview.net/profile?id=~Tianxing_Shi1}{Tianxing Shi}\quad
  \href{https://openreview.net/profile?id=~Ziyang_Huang6}{Ziyang Huang}\quad
  \href{https://openreview.net/profile?id=~Hao_Zhu5}{Hao Zhu}\quad
  \href{https://openreview.net/profile?id=~Xiao-Xiao_Long1}{Xiao-Xiao Long}\textsuperscript{\textdagger}\par}
  \vspace{3pt}
  {\small Nanjing University\quad{\footnotesize \textsuperscript{\textdagger}Corresponding author}\par}
  {\small\href{https://nju3dv-humanoidgroup.github.io/RigRecast/}{\nolinkurl{https://nju3dv-humanoidgroup.github.io/RigRecast/}}\par}}

\begin{document}
\begin{center}
{\Large\bfseries RigRecast: Large-Baseline Re-observation\\[2pt]of Dynamic Driving Scenes\par}
\PublicByline
\end{center}
\begin{minipage}{\textwidth}
\centering
\includegraphics[width=.98\textwidth]{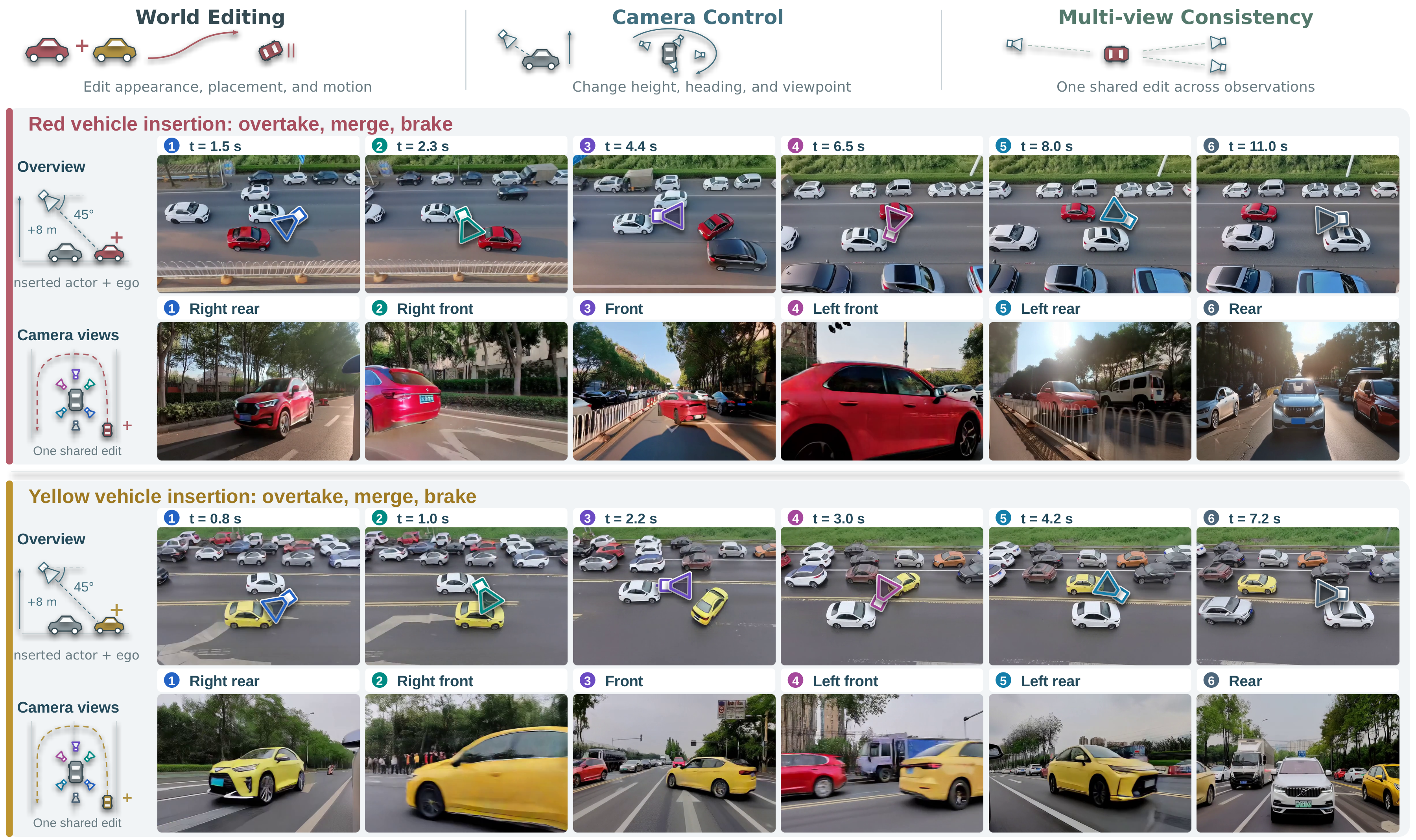}
\captionof{figure}{\textbf{One edited scene observed from multiple cameras.}
Inserted vehicles overtake, merge, and brake, shown from elevated cameras (upper rows) and rig cameras (lower rows). Columns advance in time and change the selected rig camera; vertically aligned images share a timestamp.
Within each example,
the background Gaussians, actor assets, and trajectories form one shared
world cache; CCAR refines each video independently. Editing is an
additional application; protocols are detailed in Sec.~\ref{sec:editing}.}
\label{fig:controls}
\end{minipage}

\begin{abstract}
Re-observing a recorded drive requires vehicles and pedestrians to retain
their shape, motion, and spatial relationships under new cameras.
Large viewpoint changes expose incomplete actor surfaces and distorted
background geometry that weaken the renders used to guide video generation.
We present \method{}, which completes geometry before projection by
composing vehicles and articulated humans into a shared 4D Gaussian world
from a calibrated source video, masks, and global 3D tracks and boxes
restricted to source-observed actors. Equal-volume spherical static Gaussians suppress
directional stretching in target-view renders, which guide a
Camera-Conditioned Appearance Refiner to recover appearance through source-video context and
paired camera rays, with cross-camera supervision connecting appearance
to target geometry.
On 196 held-out clips, our annotation-assisted system outperforms six reproduced
baselines on six front/side-view metrics, reaching 0.6324
Traffic IoU and 15.39\,dB PSNR. Its layout advantage extends to captured
directions approximately $178^\circ$ from the source.
Continuous camera trajectories re-observe the recorded event;
actor editing is an additional application of the same representation.
\end{abstract}

\section{Introduction}
\label{sec:introduction}
A recorded drive could be re-observed from a side camera, a taller vehicle,
or a moving virtual camera without collecting the event again.
The challenge is to change the observation without changing the event:
vehicles must reveal their other sides, pedestrians must retain their
motion, and occlusion must follow the new camera.

ReCamMaster retargets source videos
\cite{bai2025recammaster}; GEN3C renders a depth-derived 3D cache
\cite{ren2025gen3c}; TrajectoryCrafter conditions diffusion on point-cloud
renders and source video \cite{yu2025trajectorycrafter}.
Large-baseline driving synthesis further requires geometric guidance that
describes actors beyond their visible source surfaces: partial foregrounds
expose thin shells under rotation, while elongated background splats can
obscure otherwise valid actors. Refinement must then recover appearance
from conditions that omit or hide the surfaces the new camera should see.

Completing actors before projection gives new cameras geometric support
because boxes and tracks specify where actors move, but not the surfaces
and occlusions exposed by a new camera.
We introduce \method{} to compose completed vehicles, articulated humans,
and scene Gaussians into one metric world from a calibrated source video,
instance masks, and global tracks restricted to actors observed in cam0.
Source observations determine actor appearance and membership, while tracks
retain their motion after they leave the source field of view so that
target cameras can re-observe the same recorded event.

Because completed actors must remain visible in the rendering condition,
we convert static Gaussians to equal-volume spheres to suppress directional
stretching that can hide their surfaces. A Camera-Conditioned Appearance
Refiner (CCAR) then recovers appearance using source-video context, paired
camera rays, and an operation token. Cross-camera supervision trains these
interfaces against captured targets; same-camera training masks dynamic
reference regions to discourage source-layout copying. Our design couples
actor completion with static visibility correction because the completed
body is useful only if its projection reaches the refiner. These choices make
reconstructed geometry a condition for large-baseline re-observation.
World-space edits are an additional application (Fig.~\ref{fig:controls}).

On 196 held-out clips, our annotation-assisted system outperforms six
reproduced baselines across all six front/side-view metrics, with Traffic IoU of .6324 versus
.3128 for NeoVerse\cite{yang2026neoverse}, the strongest baseline on this metric.
The layout advantage holds across captured viewing directions up to
approximately $178^\circ$, spanning front, side, and rear observations.
Our contributions are:
\begin{itemize}
\item A scene-composition strategy that combines completed vehicles and articulated humans with spherical static Gaussians to construct rendering conditions for large viewpoint changes.
\item A camera-conditioned appearance refiner that combines rendered geometry, source-video context, and paired camera rays through cross-camera learning.
\item Captured-view system comparisons and adapted-component studies,
with continuous-camera and actor-editing demonstrations.
\end{itemize}

\section{Related Work}
\label{sec:related}
\paragraph{Camera-controlled video synthesis.}
ReCamMaster \cite{bai2025recammaster} learns video retargeting from
synchronized multi-camera data. GEN3C \cite{ren2025gen3c} guides generation
with a depth-derived 3D cache, while TrajectoryCrafter
\cite{yu2025trajectorycrafter} combines reprojected points and source
video in a dual-stream diffusion model. NeoVerse \cite{yang2026neoverse}
couples feed-forward 4D reconstruction with novel-trajectory generation;
AnyView \cite{vanhoorick2026anyview} learns dynamic view synthesis from
mixed 2D, 3D, and 4D supervision. ReCamDriving
\cite{li2025recamdriving} combines Gaussian guidance with progressive
camera conditioning and cross-trajectory training. Our distinction is
the construction of large-baseline rendering support: complete actor
bodies, suppress static splats that occlude them, and relate their
projections to source appearance through paired camera rays.

\paragraph{Driving reconstruction and generation.}
NeRF and Gaussian splatting \cite{mildenhall2020nerf,kerbl2023gaussians}
provide scene representations for novel-view rendering.
EmerNeRF \cite{yang2023emernerf}, Street Gaussians
\cite{yan2024street}, and DrivingGaussian \cite{zhou2023drivinggaussian}
model dynamic driving scenes; DGGT\cite{chen2026dggt} and IGRecon4D \cite{tong2026igrecon4ddynamicobjectawarefeedforwardreconstruction} predict dynamic
Gaussians and cameras in a feed-forward model.
StreetCrafter \cite{yan2025streetcrafter} conditions video diffusion on
rendered LiDAR points. We build on reconstruction but complete actors
before composition and reshape static Gaussians for generative refinement.
MagicDrive, Panacea, and DriveDreamer-2
\cite{gao2023magicdrive,wen2023panacea,zhao2024drivedreamer2} generate
multi-camera driving observations from scene controls. For long-horizon
autoregressive video generation, ENkG \cite{han2026enkg} uses the entropy
of each predicted token distribution to adapt the number of sampling candidates.
Our focus is consistency across viewpoints: each new observation depicts
the same recorded event, with cam0 providing appearance and the captured
target views providing evaluation targets.

\paragraph{Geometry-guided scene editing.}
DriveEditor \cite{liang2024driveeditor} uses 3D guidance for object
position and appearance edits; SceneCrafter \cite{zhu2025scenecrafter}
edits multi-camera driving scenes. Dream4Drive
\cite{zeng2025dream4drive}, HorizonForge \cite{wang2026horizonforge},
and SceneDirector \cite{liang2026scenedirector} connect geometry or assets
with generative models for driving interventions. HorizonForge combines
editable Gaussians and meshes with diffusion-based rendering.
DriveWeaver \cite{jiang2026driveweaver} uses point-conditioned inpainting
for vehicle insertion. In general videos, 4Director
\cite{cao2026fourdirector} moves canonical meshes with rigid transforms
and conditions generation on rendered depth. We share this
geometry-to-generation principle. Our focus is re-observing recorded
traffic under large camera changes, with articulated humans and static
covariance chosen to preserve actor visibility. Captured rig targets and
adapted-component studies evaluate this design; editing is an additional
application, not a claim of superiority over these editing systems.

\section{Method}
\label{sec:method}
\begin{figure}[tbp]\centering
\includegraphics[width=\textwidth]{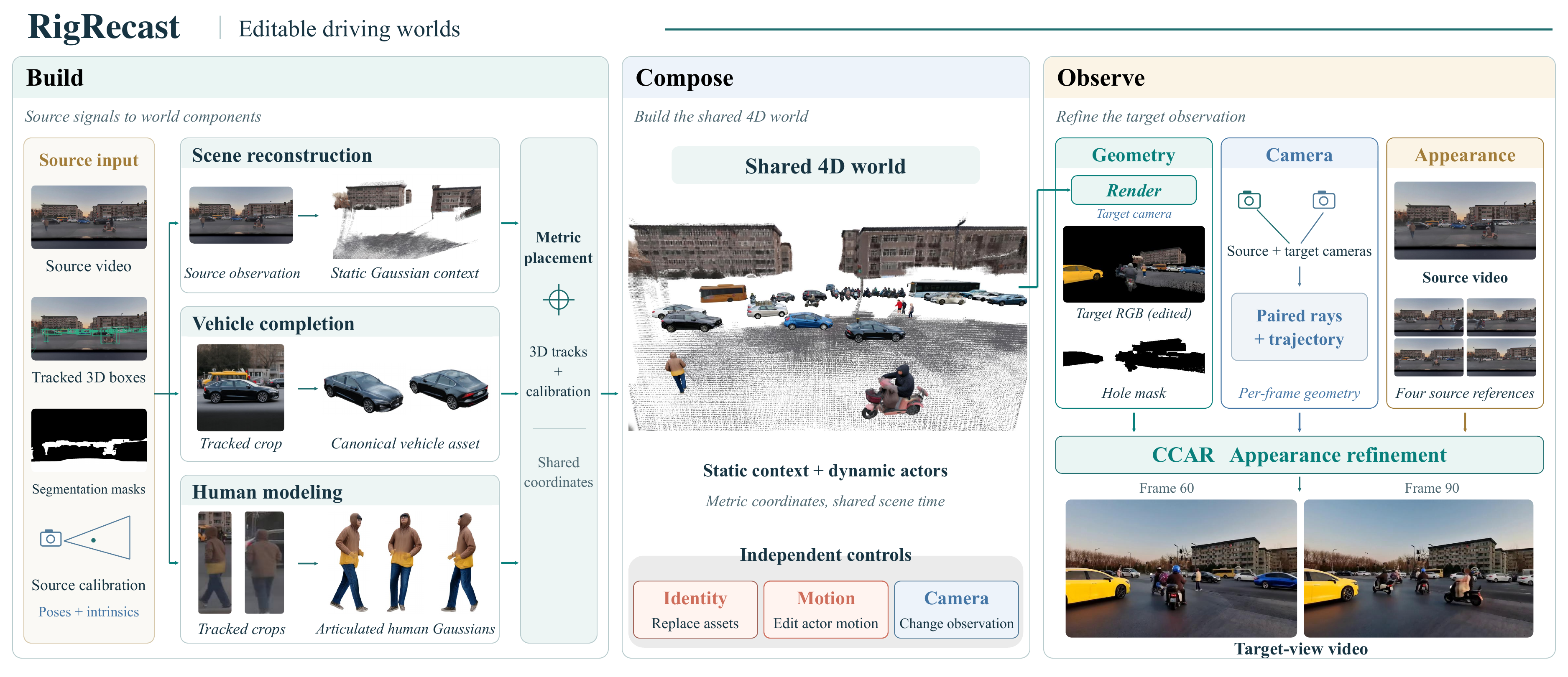}
\caption{\textbf{Complete before projection, refine afterward.}
Vehicles, humans, and scene Gaussians share metric coordinates.
CCAR combines renders and masks with source appearance, rays, and
trajectory context. The yellow-car panels share one edited sequence.}
\label{fig:pipeline}
\end{figure}

\method{} builds a shared dynamic world from a calibrated video and
actor annotations (Fig.~\ref{fig:pipeline}).
Completed actors and static scene geometry provide target-view rendering conditions; 
camera-conditioned generation combines these conditions with source appearance.

\subsection{Problem and Overview}
\label{sec:overview}
The source record is $\mathcal D^s=(\mathbf I^s,\mathcal C^s,\mathcal B,
\mathcal M)$, comprising video $\mathbf I^s=\{I_t^s\}_{t=1}^{N}$,
calibrated source cameras $\mathcal C^s$, global 3D boxes and tracks
$\mathcal B$ for source-observed actors, and instance masks $\mathcal M$.
Actors never observed in cam0 are not instantiated from annotations.
For retained actors, tracks continue outside its field of view to preserve
the recorded event rather than extrapolate unseen acceleration or turns.
A target sequence $\mathcal C^q=\{(K_t^q,T_t^q)\}_{t=1}^{N}$ specifies
intrinsics $K$ and camera-to-world poses $T\in\mathrm{SE}(3)$.
An optional edit set $\Delta$ changes actor assets or trajectories;
$\Delta=\varnothing$ retains the recorded event.

World construction $\mathcal G$ produces $\mathcal W=\mathcal G(\mathcal D^s)$.
After actor edits $\mathcal E_\Delta$, renderer $\mathcal R$ produces
RGB conditions $\mathbf G^{q,\Delta}$ and hole masks $\mathbf M^{q,\Delta}$.
The Camera-Conditioned Appearance Refiner (CCAR), $\mathcal F_\theta$,
generates the output:
\begin{align}
 (\mathbf G^{q,\Delta},\mathbf M^{q,\Delta})
 &=\mathcal R(\mathcal E_\Delta(\mathcal W),\mathcal C^q),\nonumber\\
 \widehat{\mathbf I}^{q,\Delta}
 &=\mathcal F_\theta(\mathbf G^{q,\Delta},\mathbf M^{q,\Delta};c^{s,q}).
 \label{eq:world-control}
\end{align}
Context $c^{s,q}$ contains source appearance, paired rays, and
camera-operation information. Queries change the observation; actor edits
change the world before projection. All world construction uses cam0
image evidence with the declared calibration and actor annotations.

\subsection{Completing Dynamic Participants}
\label{sec:actors}
Partial source-view surfaces omit vehicle sides and articulated body shape.
Canonical appearance assets supply these surfaces, with time-varying
placement or deformation shared across target cameras.

\paragraph{Vehicles.}
For vehicle $j$, a canonical Gaussian asset $A_j$ supplies body shape,
while $H_{j,t}$ places it along the retained track. Boxes initialize
metric placement and extent; per-asset source-view alignment refines the
fit before rendering for training or testing. The benchmark uses Asset
Harvester assets completed from cam0 crops \cite{cao2026assetharvester}.
Target RGB is not used for fitting. Cameras reproject the same asset;
the supplement details alternative completion routes.

\paragraph{Pedestrians.}
Human motion additionally requires pose-dependent deformation.
GVHMR \cite{shen2024gvhmr} estimates pose $q_{k,t}$ from associated cam0
crops, and LHM++ \cite{lhmplusplusRepository} supplies a canonical human
asset $B_k$. Calibrated rays, box depth, and height alignment determine
world placement $H^p_{k,t}$. With articulated deformation $D$, the posed
Gaussians are
\begin{equation}
 P_k(t)=H^p_{k,t}D(B_k,q_{k,t}).\label{eq:human}
\end{equation}
Appearance follows the canonical body through deformation; cameras at
the same scene time observe the same posed geometry.

\subsection{View-Robust World Composition}
\label{sec:world}
Background geometry must not obscure completed actors under camera changes.
Our DGGT-based reconstruction
\cite{chen2026dggt} uses depth and pose alignment to place scene Gaussians
in metric coordinates, with InfiniDepth-assisted metric reconstruction
\cite{yu2026infinidepth} in the archived scene-building route.
Anisotropic static splats can fit source appearance yet stretch across
novel views. We replace their directional covariance with an equal-volume
sphere:
\begin{equation}
 r_i=(\det\Sigma_i)^{1/6},\qquad
 \Sigma_i^{\rm sph}=r_i^2 I_3.\label{eq:isotropic}
\end{equation}
Here $\Sigma_i$ is the original covariance, $I_3$ the identity, and $r_i$
the geometric mean of its principal standard deviations.
This preserves volume at a fixed Mahalanobis radius while removing
directional elongation. Centers, colors, opacities, and dynamic asset
covariances remain unchanged. The condition prioritizes geometric support;
CCAR recovers missing appearance.

We remove completed actors' original foreground points, resolve heading
ambiguities, and place assets along smoothed tracks.
Let $S$ contain static spherical Gaussians and
$U_t$ the residual dynamic points without explicit replacement. Then
\begin{equation}
 \mathcal W_t=S\cup U_t\cup
 \bigcup_{j\in\mathcal V_t}H_{j,t}A_j\cup
 \bigcup_{k\in\mathcal P_t}P_k(t),\label{eq:composition}
\end{equation}
where $\mathcal V_t$ and $\mathcal P_t$ index vehicles and pedestrians.
Asset edits modify $A_j$ or the actor set; motion edits modify $H_{j,t}$,
leaving other actor states unchanged.
Rendering resolves visibility jointly. Rig controls update camera centers
and orientations through mounting extrinsics and ego motion, preserving
the actor timeline.

\subsection{Camera-Conditioned Appearance Refinement}
\label{sec:ccar}
Even with completed actors, holes and appearance errors leave rendered
observations unsuitable as final videos.
CCAR adapts Wan2.1-VACE-14B \cite{wan2025,jiang2025vace}:
target renders and hole masks enter the VACE control branch alongside
four source references $Q_k^s$, $k=1,\ldots,4$.
A source-video adapter adds temporal appearance context beyond these
four references to DiT.

\paragraph{Paired camera geometry.}
To relate source appearance to target geometry, we encode both camera
sequences in the first source-camera frame \cite{he2024cameractrl},
with intrinsics adjusted to the actual crop and resize.
At image location $u$, let $d_t^v(u)$ be a unit ray direction and $o_t^v$
the camera center normalized by 20\,m, for $v\in\{s,q\}$.
The paired Pl\"ucker representation is
\begin{equation}
 \pi_t^v(u)=[d_t^v(u),o_t^v\times d_t^v(u)],\quad
 \Pi_t^{s,q}=[\pi_t^s,\pi_t^q].\label{eq:rays}
\end{equation}
Directions encode viewing orientation; moments carry origin-dependent
spatial information.
We sample pairs at latent-grid centers: reference slots use matching
source rays in both halves, while video slots use source/target rays
at the matching latent time.

\paragraph{Conditioning interfaces.}
Let $h_D,h_V$ be the DiT and VACE patch embeddings and $z^s$ the encoded
source sequence. Source adapter $A_s$ and separate ray adapters $A_D,A_V$
modify these embeddings as
\begin{align}
 h_D' &= h_D+A_s(z^s)+A_D(\Pi^{s,q}),\nonumber\\
 h_V' &= h_V+A_V(\Pi^{s,q}).\label{eq:ccar-injection}
\end{align}
Thus source appearance enters the video branch, while both branches
receive the camera relationship. Adapter output projections start at
zero. A separate MLP maps a 17-dimensional descriptor of the operation,
relative pose, camera ID, and mirroring to one text-context token.
This token summarizes the operation; dense rays retain its per-frame
camera geometry. CCAR parameters are shared across target cameras, 
while each video is refined independently from the common world representation.

\subsection{Training and Inference}
\label{sec:training}
Synchronized camera pairs teach refinement across different observations
of one event. Cam0 supplies the source; cam0--6 supply targets.
The source-built world provides renders and geometry, while captured
target RGB provides supervision. Target images from this disjoint training
split never enter test-world construction or inference conditions.

\paragraph{Dynamic reference masking.}
Same-camera reconstruction can encourage copying reference actors over
rendered geometry. For target camera index $c_q$, one masking decision
governs all four references:
\begin{align}
 b&\sim\operatorname{Bernoulli}(0.5\,\mathbf1[c_q=0]),\nonumber\\
 \widetilde Q_k^s&=\operatorname{MaskDyn}(Q_k^s,bM_k^{\rm dyn}),
 \quad k=1,\ldots,4.\label{eq:refmask}
\end{align}
Here $M_k^{\rm dyn}$ is the dynamic-region union in reference $k$.
$\operatorname{MaskDyn}$ marks these pixels unknown and gates their
encoded reference content, rather than simply painting them black.
Cam0 targets jointly mask dynamic regions with 50\% probability;
other targets retain references. Slots, rays, and the complete source
video remain present.

\paragraph{Geometry-weighted adaptation.}
We emphasize actors and missing render regions during adaptation.
Following flow matching \cite{lipman2022flow},
we interpolate clean target latent $z_0$ and Gaussian noise $\epsilon$
at noise level $\sigma$:
\begin{equation}
 z_\sigma=(1-\sigma)z_0+\sigma\epsilon,\qquad
 v^*=\epsilon-z_0.\label{eq:flow}
\end{equation}
Let $m_h,m_a,m_v$ denote hole coverage, actor coverage, and non-ego
validity on the latent grid. With $w_h=w_a=2$, the objective is
\begin{align}
 w&=m_v[(1-m_a)(1+(w_h-1)m_h)+w_a m_a],\nonumber\\
 \mathcal L&=\mathbb E\!\left[\lambda(\sigma)
 \frac{\sum_{n,x}w_{n,x}\|v_\theta-v^*\|_2^2}
 {\max(C_z\sum_{n,x}w_{n,x},1)}\right].\label{eq:loss}
\end{align}
Here $v_\theta$ is the velocity predicted from $z_\sigma$, $\sigma$, and
the packed conditions; $n,x$ index latent time and position, $C_z$ is
the channel count, and $\lambda$ is the noise-schedule weight.
Actor coverage overrides hole weighting rather than multiplying it.
Ego regions and reference-prefix positions do not contribute to the loss.
We freeze the pretrained DiT/VACE, VAE, and text encoder, and train the
added adapters and attention/FFN LoRA updates \cite{hu2021lora}.
VACE warmup precedes joint adaptation; the supplement specifies packing,
ranks, and optimization settings.

\paragraph{Inference.}
We render the edited world at each target camera, using matching
parameters for its rays and operation token. Camera-only queries retain
source appearance.
Standard cam0 editing masks edited-vehicle reference regions; other
cameras retain original references. The supplement documents
demonstration-specific compositing.

\section{Experiments}
\label{sec:experiments}
We evaluate captured-view fidelity, diagnose geometry and conditioning
through component adaptation, and test edits across camera observations.

\subsection{Experimental Setup}
\label{sec:setup}
\paragraph{Data and tasks.}
The held-out industrial test set contains 196 synchronized multi-camera
clips spanning daytime, nighttime, urban roads, and highways, disjoint
from training under the provider's split audit. Cam0 supplies source RGB;
cam1--6 provide captured targets. We score the common first 81 frame IDs
at $704\times480$. The primary comparison uses cam1/3/4, and the full-rig
evaluation adds cam2/5/6. Their optical-axis differences from cam0 are
approximately $1^\circ$, $70^\circ$, $140^\circ$, and $178^\circ$ for
cam1, cam3/4, cam2/5, and cam6. Component studies use 20 stratified test clips.

\paragraph{Inputs and comparisons.}
World construction uses cam0 RGB, calibration, masks, and global
tracks/boxes only for actors observed in cam0. Their tracks continue
beyond source visibility; never-observed actors are not added from boxes.
Target cameras define queries; target RGB is reserved for evaluation.
We reproduce GEN3C, NeoVerse, AnyView, TrajectoryCrafter, ReCamMaster,
and ReCamDriving on matched scenes and timestamps
\cite{ren2025gen3c,yang2026neoverse,vanhoorick2026anyview,
yu2025trajectorycrafter,bai2025recammaster,li2025recamdriving}.
This system-level comparison retains different input-information budgets
and training histories: continued actor tracks provide extra geometry,
so the gains do not establish superiority under matched geometric inputs.
Separately, VACE + LoRA receives Full CCAR's identical completed renders,
providing a render-matched refinement baseline (Table~\ref{tab:components}).
The supplement details temporal packing
and camera adaptations. Paired benchmarks use step5250; camera/editing
demos use step8000. Component adaptation is specified below.

\paragraph{Metrics.}
Traffic IoU measures vehicle, pedestrian, and two-wheeler layout with
independent YOLOv8x/\allowbreak SAM2 segmentation of generated and captured videos
\cite{ultralytics2023yolov8,ravi2024sam2}. Target ego masks, expanded
upward by ten pixels, exclude the same region on both sides. Intersections
and unions accumulate within clips; empty unions and segmentation failures
follow a common exclusion rule. Traffic IoU, PSNR, SSIM
\cite{wang2004ssim}, and LPIPS \cite{zhang2018lpips} average clips within
physical scenes, then scenes equally. FID \cite{heusel2017fid} and FVD
\cite{unterthiner2018fvd} pool matched features with common preprocessing.

\subsection{Large-Viewpoint Video Generation}
\label{sec:nvs}
\begin{figure}[tbp]\centering
\includegraphics[width=\textwidth]{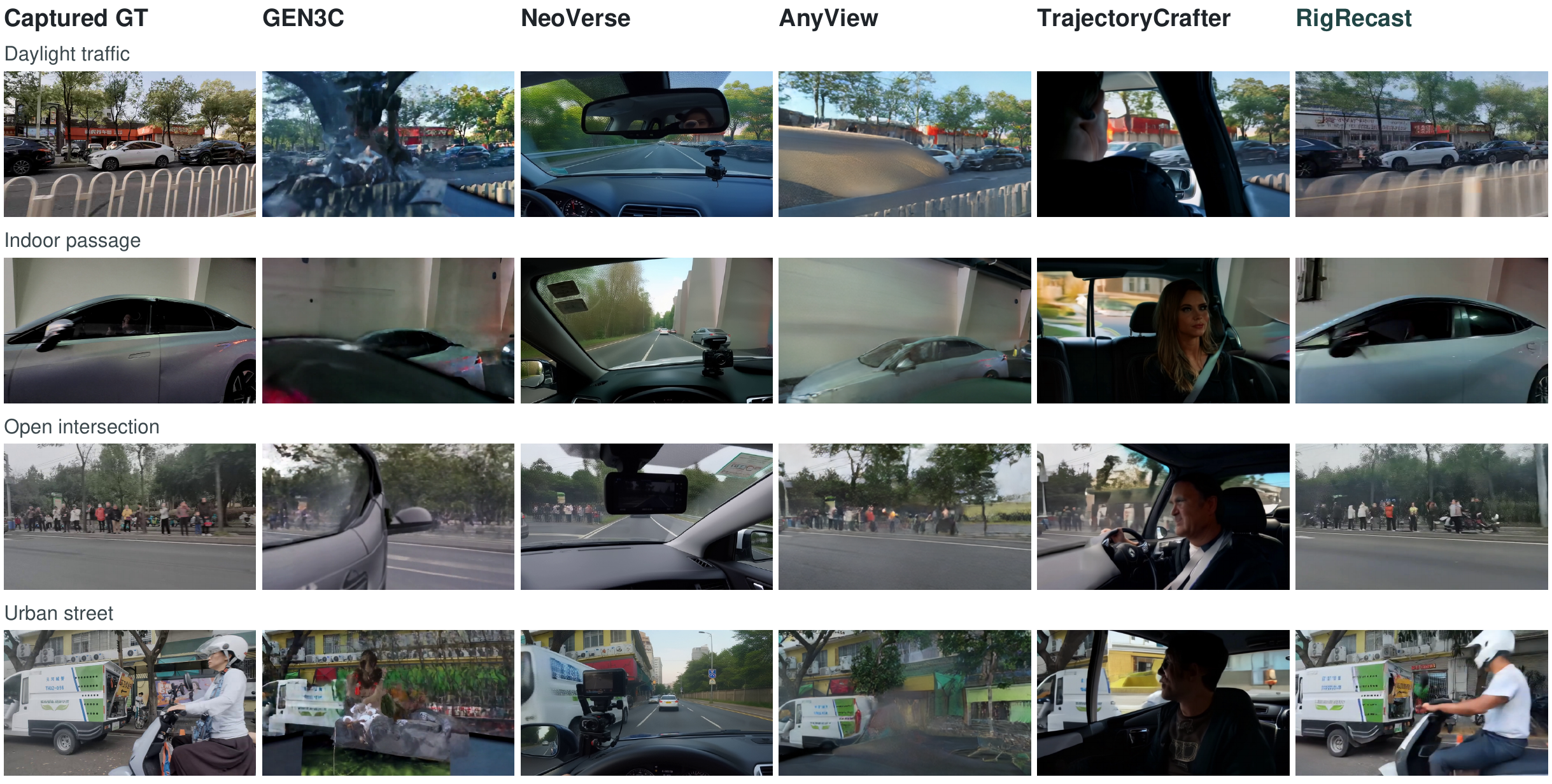}
\caption{\textbf{Recovering the requested view and traffic event.}
Each row compares one timestamp across methods. From top to bottom:
cars passing behind a roadside barrier, a passing car in an indoor passage,
pedestrians waiting at an intersection, and a rider on an urban
street. The captured target at left anchors the comparison;
look across each row for actor position, extent, and occlusion.}
\label{fig:peers}
\end{figure}

\paragraph{Recovery of captured target views.}
\method{} leads all six metrics on front and side targets
(Table~\ref{tab:main}). Traffic IoU improves from NeoVerse's .3128 to
.6324, while PSNR exceeds the strongest baseline by 2.52\,dB.
Scene-bootstrap 95\% intervals for these gains are $[.2655,.3704]$ and
$[1.882,3.115]$\,dB, respectively. Figure~\ref{fig:peers} highlights recovery of
the requested camera and traffic layout, beyond image sharpness.

\begin{table}[tbp]\centering
\caption{\textbf{Improved layout and appearance on captured views.}
Cam1/3/4 over 196 held-out clips, scoring the same first 81 frame IDs.
Best values are bold; method-specific input/FOV adaptations are retained.}
\label{tab:main}
\begingroup\small\setlength{\tabcolsep}{4pt}\begin{tabularx}{\linewidth}{@{}Xrrrrrr@{}}\toprule
\metrichead\\\midrule
GEN3C & .1040 & 11.79 & .2503 & .7350 & 183.67 & 1611.35\\
NeoVerse (source $K$) & .3128 & 12.21 & .3132 & .6859 & 133.34 & 1773.56\\
AnyView ($3\times41$) & .1502 & 12.71 & .3934 & .7165 & 128.27 & 1140.81\\
TrajectoryCrafter ($3\times49$) & .1656 & 10.52 & .2215 & .7499 & 157.76 & 2150.92\\
ReCamMaster$^{\dagger}$ & .1327 & 12.87 & .4109 & .7595 & 154.71 & 1214.18\\
ReCamDriving$^{\ddagger}$ & .1960 & 11.91 & .3401 & .6864 & 117.10 & 1322.23\\
\method{} & \textbf{.6324} & \textbf{15.39} & \textbf{.4644} & \textbf{.4956} & \textbf{46.71} & \textbf{537.22}\\
\bottomrule\end{tabularx}
\par\smallskip{\footnotesize
$\dagger$ Source-FOV adaptation for cam1; native source FOV for other cameras.
$\ddagger$ Real-rig extension using source-$K$ geometry.
All image metrics use equal physical-scene aggregation; FID/FVD use pooled features.}
\endgroup
\end{table}

\paragraph{Side-rear and backward observations.}
\method{} leads Traffic IoU in every direction group (Table~\ref{tab:angles}), with
.5381 IoU near $140^\circ$ and .3775 near $178^\circ$. All-rig scores
reach .5585 IoU and 14.42\,dB PSNR. Rig transfers include mounting
translation and FOV changes, not isolated yaw; cam1 tests telephoto FOV.

\begin{table}[tbp]\centering
\caption{\textbf{Layout gains extend to backward observations.}
Traffic IoU grouped by approximate optical-axis difference:
cam1, cam3/4, cam2/5, and cam6. The protocol matches
Table~\ref{tab:main}; values below .00005 round to .0000.}
\label{tab:angles}
\begingroup\small\setlength{\tabcolsep}{4pt}\begin{tabularx}{\linewidth}{@{}Xrrrr@{}}\toprule
Method & $\sim1^\circ$ & $\sim70^\circ$ & $\sim140^\circ$ & $\sim178^\circ$\\\midrule
GEN3C & .0320 & .1400 & .0000 & .0000\\
NeoVerse & .5921 & .1731 & .1516 & .1380\\
AnyView & .1270 & .1618 & .0710 & .1012\\
TrajectoryCrafter & .1357 & .1805 & .1186 & .1133\\
ReCamMaster & .2523 & .0728 & .0743 & .0948\\
ReCamDriving & .2673 & .1604 & .1331 & .1389\\
\method{} & \textbf{.7236} & \textbf{.5869} & \textbf{.5381} & \textbf{.3775}\\
\bottomrule\end{tabularx}
\endgroup
\end{table}

\begin{figure}[tbp]\centering
\includegraphics[width=\textwidth]{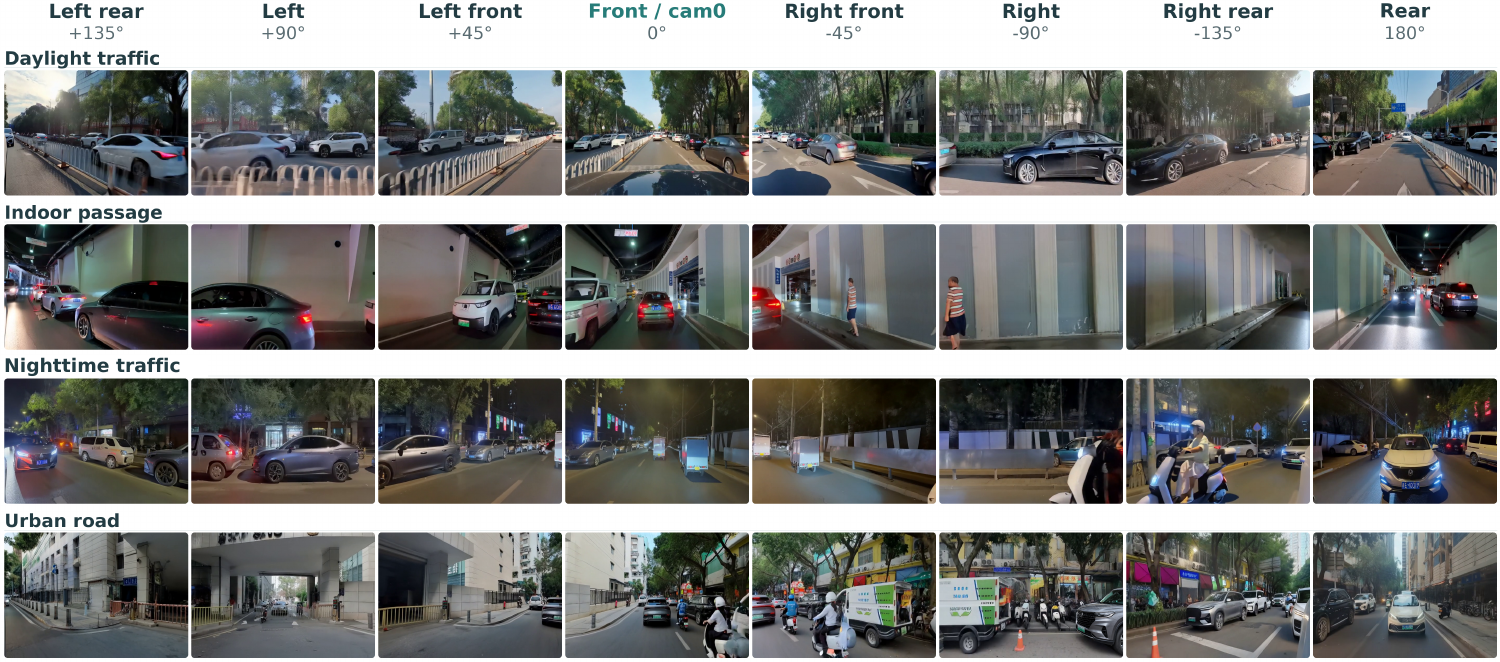}
\caption{\textbf{Shared events across eight camera headings.}
Each row shows one scene at offset 60 (6\,s), at one optical center
with headings spaced by $45^\circ$; front aligns with cam0.
Overlapping views expose complementary actor surfaces and occlusions.}
\label{fig:synchronized-views}
\end{figure}

\paragraph{Continuous and synchronized camera controls.}
Figure~\ref{fig:synchronized-views} shows recognizable actors across
eight headings in four scenes, sharing geometry and time with per-video
CCAR. Supplementary sequences extend control to translation, elevation,
virtual trucks, $90^\circ$ rig drift, and $360^\circ$ rotation. Drift
rotates mounting centers and orientations about the moving ego origin,
without changing actor tracks. Paired accuracy is measured only where
captured targets exist.

\subsection{Ablation Studies}
\label{sec:ablations}
We adapt each component intervention from step8000 for 500 updates at
learning rate $10^{-4}$ on 98 stratified training clips, excluding test
scenes. Evaluation uses 20 separate clips, seed 42, 30 sampling steps,
CFG 5, and control scale 1.0, scoring the first 81 of 121 generated frames.
Full CCAR receives the same additional budget; variant-specific training
settings are detailed in the supplement.

\begin{table}[tbp]\centering
\caption{\textbf{Geometry and learned camera conditions both contribute.}
Twenty test clips, cam1/3/4, first 81 frames. Except for Native VACE,
all rows use 500-step adaptation from step8000 on 98 disjoint training
clips at learning rate $10^{-4}$. Best values are bold.}
\label{tab:components}
\begingroup\small\setlength{\tabcolsep}{4pt}\begin{tabularx}{\linewidth}{@{}Xlrrrr@{}}
\toprule
Configuration & Change & IoU $\uparrow$ & PSNR $\uparrow$ & SSIM $\uparrow$ & LPIPS $\downarrow$\\
\midrule
Native VACE & Pretrained, no adaptation & .4485 & 11.35 & .2419 & .8222\\
VACE + LoRA (same render) & No added conditioning modules & .4612 & 12.02 & .2735 & .7930\\\midrule
w/o actor completion & No completed actors & .3215 & 10.31 & .1937 & .7502\\
w/o spherical static & Anisotropic static covariance & .4046 & 11.19 & .3032 & .6754\\\midrule
w/o dense rays & No ray adapters & .5763 & 12.91 & .3766 & .6313\\
w/o trajectory token & No operation token & .5517 & 13.28 & .3512 & .6701\\
w/o source-video adapter & No full-video appearance adapter & .5931 & 13.25 & .3540 & .6680\\\midrule
Full \method{} & All components & \textbf{.6156} & \textbf{14.86} & \textbf{.4376} & \textbf{.5209}\\
\bottomrule
\end{tabularx}

\endgroup
\end{table}

\begin{figure}[tbp]\centering
\includegraphics[width=.98\textwidth]{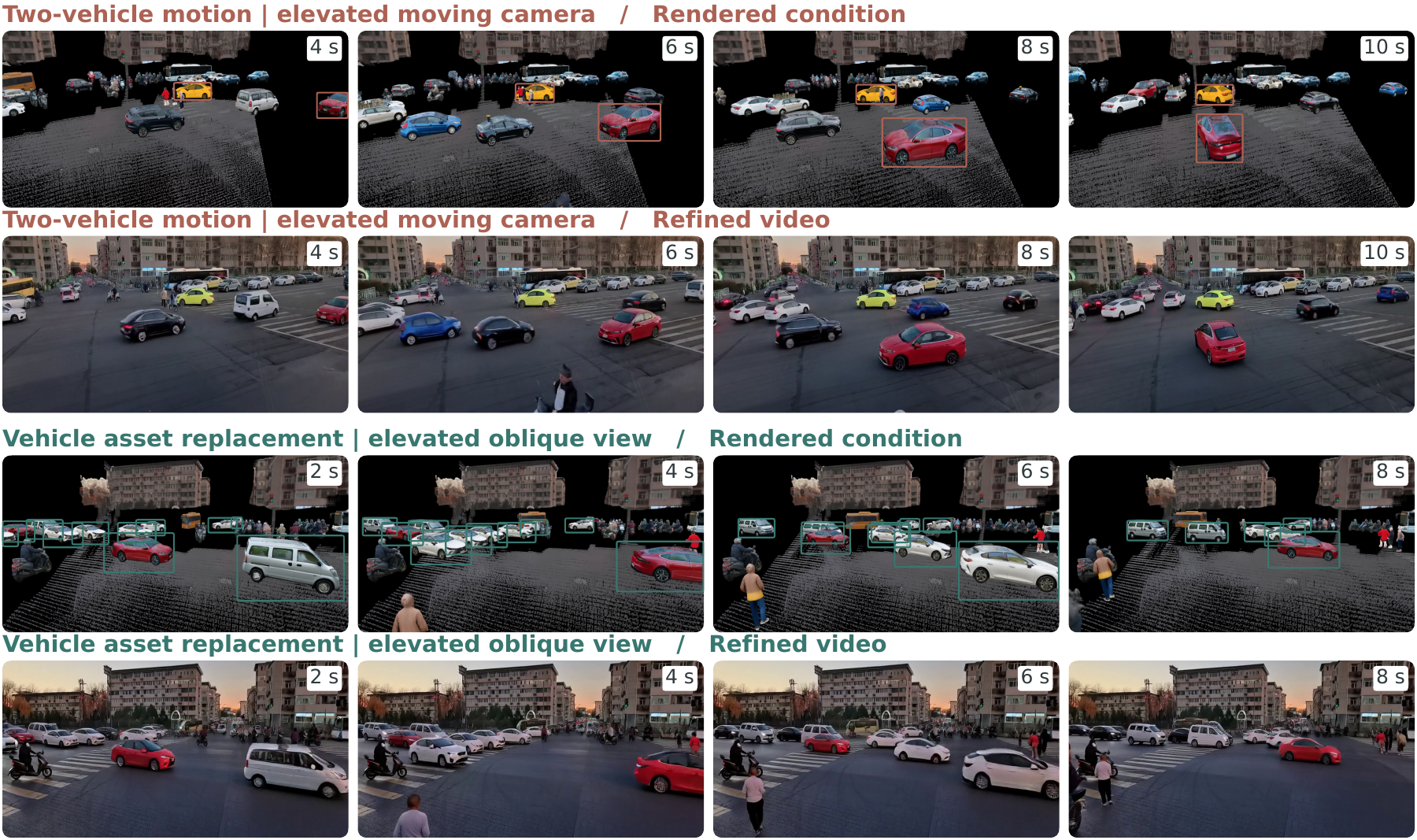}
\caption{\textbf{Actor controls under different camera observations.}
Top pair (4/6/8/10\,s): two vehicles are inserted into the same scene.
The red car enters from the right and merges while the yellow car
advances ahead, seen from a moving camera raised by 3\,m.
Bottom pair (2/4/6/8\,s): vehicle assets are replaced on their original
tracks, seen from an elevated oblique camera. Each pair shows the render
above the refined output; boxes mark edited actors in the renders only.}
\label{fig:edit-pair}
\end{figure}

\paragraph{A render-matched strong baseline.}
VACE + LoRA receives Full CCAR's identical completed renders, including
static context and actors placed using the same global boxes, tracks,
and calibration. It already benefits from our annotation-assisted geometry.
Under the matched adaptation protocol above, Full improves IoU from
.4612 to .6156 on 20 test clips.
The gain reflects the added source-video, dense-ray, and trajectory
conditioning, not extra rendered geometry.

\paragraph{Geometric support after adaptation.}
The w/o actor completion variant retains raw DGGT foregrounds and reaches
.3215 IoU after adaptation, versus .6156 for Full CCAR. Native VACE receives
completed renders and reaches .4485 without adaptation. These results
compare the evaluated foreground-processing recipes
(Table~\ref{tab:components}).
Source-facing fragments can reproduce an observed surface but offer
little support for newly exposed sides; completed assets provide that
support before refinement.
Static visibility is equally important: anisotropic context lowers IoU
from .6156 to .4046. Stretched
splats occlude actors; apparently raised ground hides vehicle undersides
and pedestrians' lower bodies. Sphericalization reduces this false
occlusion and exposes more of the actors at their unchanged placements.
The resulting improvement concerns visible geometric support, with
actor positions held fixed.

\paragraph{Camera conditions under sparse render support.}
Removing dense rays or the trajectory token reduces
IoU from .6156 to .5763 or .5517. Qualitative side-rear sequences show
incorrect views when slow ego motion leaves rear render support sparse.
Without the token, apparent speed changes and large unintended turns
also occur. Dense rays remain in that variant, suggesting that the
operation summary complements per-frame geometry when render support is weak.

\paragraph{Appearance context and observation fidelity.}
Removing the source-video adapter weakens illumination and texture,
raising LPIPS from .5209 to .6680. Four references and camera conditions
remain, but full-window appearance context is absent.
Full-video context supplies appearance changes between reference times.

\subsection{Additional Application: Actor Editing}
\label{sec:editing}
\paragraph{Shared controls under different observations.}
Asset and trajectory edits modify one world before per-camera refinement.
Standard cam0 editing masks edited vehicles in the references; other
cameras retain complete references. Figure~\ref{fig:edit-pair} shows
insertion and replacement with spherical static context.
Counterfactual outputs are assessed against the requested edit, not
against unedited captured GT.

Figure~\ref{fig:controls} additionally shows overtaking, merging, and
braking. Each example shares its background Gaussians, actors, and tracks
across rig and elevated observations, with the supplement's demo-specific
references. These applications are separate from paired evaluation.

\paragraph{Adherence depends on the requested operation.}
A ten-scene, 84-request pilot uses Qwen3-VL-32B with limited human
adjudication. Standard edits attain 95.9\% strict success; stress edits
attain 19.6\%. Overall, 116/210 judgeable pairs succeed (55.2\%), with two
unjudgeable. The supplement reports operation-wise and joint-view results;
this pilot characterizes an application rather than comparative editing superiority.

\FloatBarrier
\section{Conclusion}
\method{} completes dynamic geometry before camera projection and
recovers appearance afterward. Canonical actors and spherical static
context provide visible geometric support, while CCAR connects it to
source appearance and the requested camera. Captured-view comparisons
show gains across front, side, and rear cameras; component adaptation
links different failure modes to geometric support, camera conditioning,
and temporal appearance. Continuous controls and actor edits reuse this
world instead of reconstructing each observation.
For camera-only queries, recorded actor tracks remain fixed while the
observer's position, orientation, and intrinsics can change.

\textbf{Observation coverage remains the principal constraint.}
Short recordings or slow ego motion limit reconstructed support for
rear observations. Additional calibrated cameras have alleviated this
in a separate multi-source setting. Our single-RGB-source benchmark
also relies on calibration and continued tracks of source-observed actors.
Independent refinement may alter actor identity or soften unusual poses.
Shared world geometry therefore supports control without guaranteeing
every generated appearance detail.

\section*{LLM Usage Statement}
ChatGPT/Codex assisted with literature discovery, translation, manuscript
organization, LaTeX preparation, and experimental implementation, execution,
and analysis. The authors verified the scientific content, citations,
reported measurements, and their underlying artifacts. Qwen3-VL-32B is
also used as an evaluation agent for the ten-scene editing pilot, with
limited human adjudication as described in the supplementary protocol.
Generative models
in the method produce assets and synthesized videos; benchmark ground truth
consists of captured observations.

\clearpage
\appendix
\section*{Supplementary material}
\phantomsection\addcontentsline{toc}{section}{Supplementary material}
\setcounter{figure}{0}\setcounter{table}{0}
\renewcommand{\thefigure}{S\arabic{figure}}\renewcommand{\thetable}{S\arabic{table}}
\renewcommand{\theHfigure}{S.\arabic{figure}}\renewcommand{\theHtable}{S.\arabic{table}}
\section{World and Camera Conventions}
\textbf{Every control is evaluated in a common metric world and timeline.}
Camera poses are camera-to-world transforms; projection uses their
inverses. Source and target ray coordinates are expressed relative to the
first source camera. Reconstruction scale is converted to meters before
the fixed 20\,m ray-origin normalization. Intrinsics follow the exact
resize and crop. In the fresh component study, source images are
isotropically filled/cropped to $832\times480$ and then resized to
$704\times480$; rendering and rays use the resulting effective intrinsics.

\begin{figure}[tbp]\centering
\includegraphics[width=\textwidth]{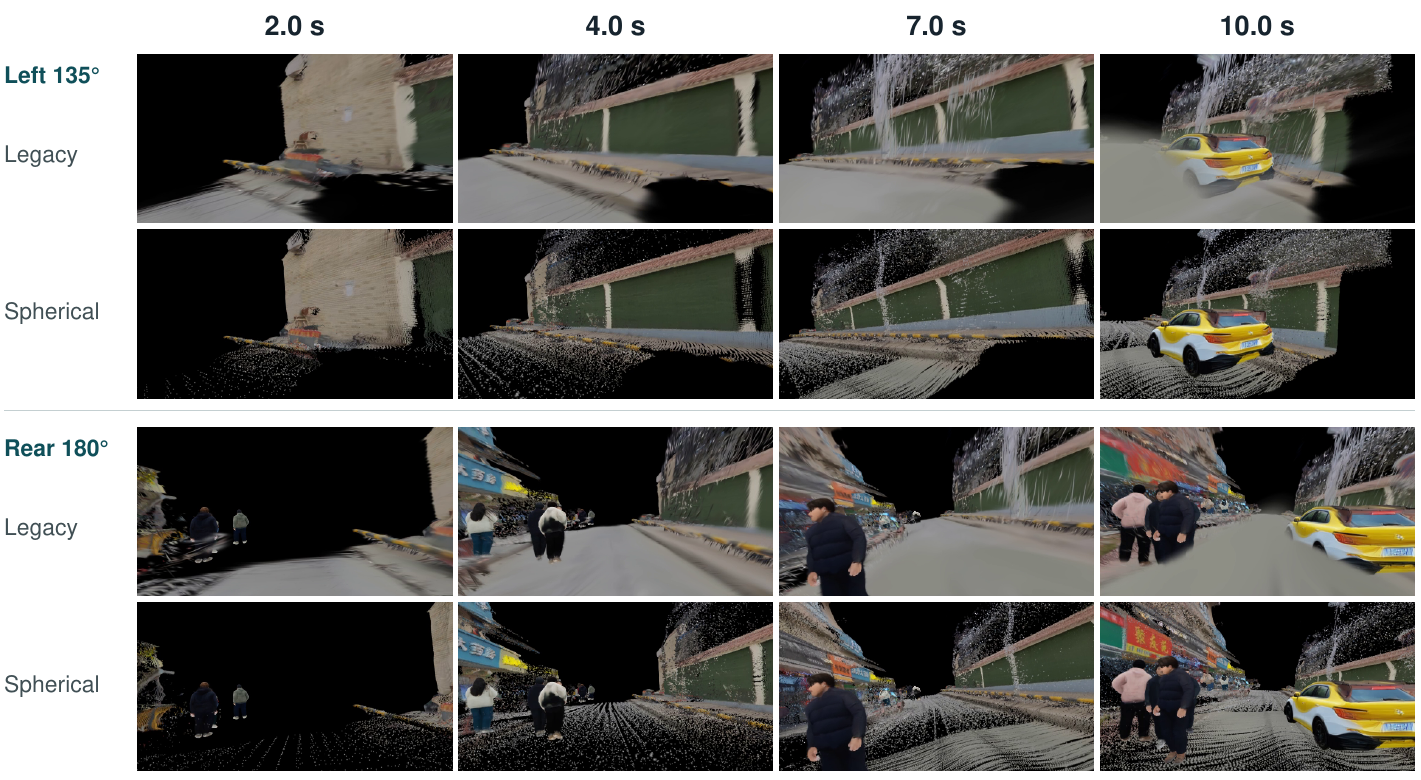}
\caption{\textbf{Looking back exposes stretched static geometry.}
Columns follow scene 3694 at frames 20, 40, 70, and 100 (10\,Hz).
For each heading (left $135^\circ$, rear $180^\circ$), the upper row
shows the legacy render and the lower row the spherical static Gaussians
at the same time and crop. Along the wall and road, stretched surfaces
give way to gaps between samples. At 10\,s the difference is also
visible below the passing yellow vehicle. These are geometry renders;
Video S01 shows the full sequence.}
\label{fig:supp-spherical}
\end{figure}

\textbf{Canonical assets preserve the distinction between shape and
placement.} Box fitting rescales asset centers along object axes and
updates Gaussian scales using the fitting transform's axis magnitudes.
World placement applies actor rotation, translation, and uniform scale;
Gaussian orientations rotate with the actor. The main text's $H_{j,t}$
includes this fitting and placement. Completed tracks replace their
original foreground points; unreplaced dynamic points remain in $U_t$.
Static equal-volume sphericalization leaves all dynamic-asset covariance
parameters unchanged.
Figure~\ref{fig:supp-spherical} shows where the choice of static support
becomes visible when looking back along the road.

\textbf{Rig edits move camera centers as well as viewing directions.}
The left-turn drift rotates the rig from $0$ to $90^\circ$ over offsets
20--40, then holds its orientation while the ego origin continues forward.
Virtual truck views retain native mounting heights and intrinsics; the
mean-overlap alignment uses a longitudinal offset of $-3.285$\,m with
unchanged rig yaw. All cameras share scene timestamps and actor motion.

\section{Refinement and Training Details}
\begin{figure}[tbp]\centering
\includegraphics[width=\textwidth]{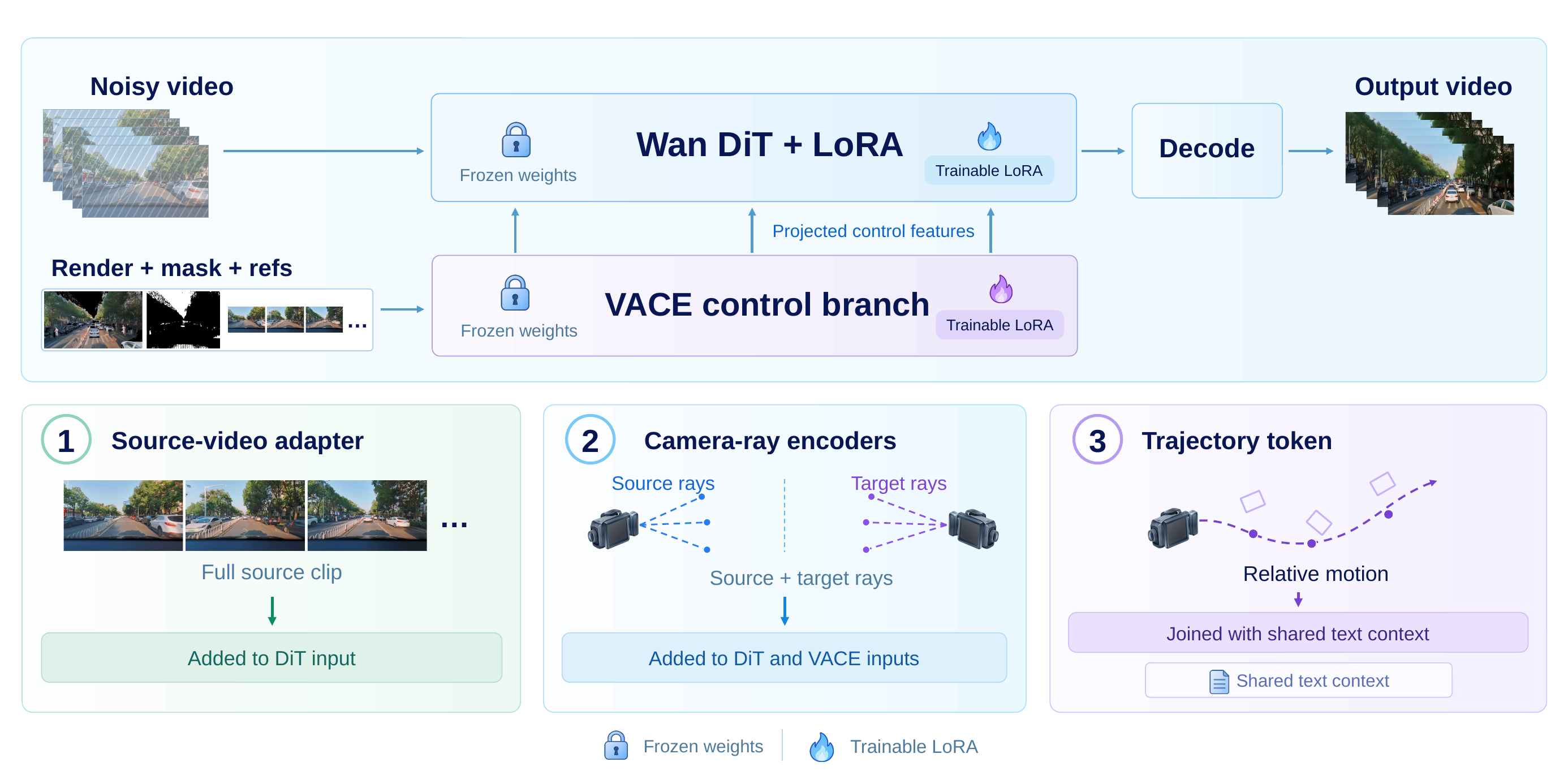}
\caption{\textbf{CCAR separates appearance, geometry, and operation context.}
Top: the noisy target latent is denoised by Wan DiT with projected VACE
control features. Bottom: the source-video adapter enters DiT, paired-ray
encoders enter both branches, and the trajectory token joins text context.
Pretrained weights are frozen; the three conditioning modules and
DiT/VACE LoRA are trainable.}
\label{fig:ccar-detail}
\end{figure}

Figure~\ref{fig:ccar-detail} shows how appearance, camera geometry, and
operation context enter the two pretrained branches.
\subsection{Condition Packing}
\textbf{Reference appearance and video-time geometry have matching latent
positions.} The VAE compresses a 121-frame sequence to 31 temporal positions.
Four references are encoded independently and prepended, yielding 35
positions. Their paired-ray entries use source rays at offsets
0, 40, 80, and 120 in both halves of the 12-channel tensor. Video positions
use source/target rays at offsets $0,4,\ldots,120$. Spatial ray samples are
latent-grid centers mapped back to the image through the effective intrinsics.

\textbf{Masking withholds dynamic reference content without removing its
camera slot.} For cam0 targets, one Bernoulli draw with probability .5
jointly masks dynamic regions in all four reference frames. The availability
mask marks these regions unknown and the encoding path gates their reference
latents. Matching ray slots remain present, as does the complete source
video. Standard cam0 editing always masks edited-vehicle regions in the
references; other cameras retain original references. Demonstration-specific
compositing and the earlier fixed-reference diagnostic are separately
documented in Sec.~\ref{sec:edit-protocol}.

\textbf{Separate adapters inject appearance, geometry, and operation
context into the pretrained refiner.} The source adapter consumes 16-channel
VAE latents with four reference-prefix positions. The VACE input packs
16-channel known-region render latents, 16-channel hole-region latents,
and 64 mask channels. Each $8\times8$ image-mask block is packed into
channels and temporally resampled. Reference positions contain reference
latents and their availability masks. DiT and VACE have separate
12-channel ray encoders. Each spatial-temporal adapter uses a
$3\times3\times3$ convolution, 128 hidden channels, 32-group normalization,
SiLU, and a projection with kernel and stride $(1,2,2)$ to 5120 channels.
The projection starts at zero and has a learned scalar gain initialized
to one. Source features enter DiT; ray features enter both patch embeddings.

\textbf{The learned additions meet at token-level interfaces.}
If $h_D$ and $h_V$ denote the pretrained DiT and VACE patch embeddings,
the conditioned embeddings are
\begin{align}
 h_D' &= h_D + A_s(z^s) + A_D(\Pi^{s,q}),\\
 h_V' &= h_V + A_V(\Pi^{s,q}).
\end{align}
Here $z^s$ is the encoded source sequence and $A_s,A_D,A_V$ are the
source-video and two ray adapters. The trajectory MLP supplies a separate
text-context token. Dense spatial control, temporal appearance context,
and operation-level conditioning augment the pretrained denoiser.

\subsection{Trajectory Context}
\textbf{The trajectory token summarizes an operation while dense rays
retain its full camera sequence.} The checkpoint's 17 entries comprise
five operation flags, normalized yaw/pitch/roll, sine and cosine of yaw,
three translations, translation and angle-summary magnitudes, camera ID,
and a mirror flag (Table~\ref{tab:trajectory}). The MLP maps 17 inputs
through 1024 hidden units and SiLU to a 5120-dimensional token appended
after text embedding, with zero-initialized output and no dropout.

\begin{table}[tbp]\centering\small
\caption{Ordered trajectory features. Angles are in degrees and
translations in meters.}\label{tab:trajectory}
\begin{tabularx}{\linewidth}{@{}lX@{}}\toprule
Entries & Contents\\\midrule
1--5 & Identity, camera transfer, translation, yaw, combined-operation flags\\
6--8 & Yaw/180, sine of yaw, cosine of yaw\\
9--10 & Pitch/180, roll/180\\
11--13 & Translation components divided by 20\\
14--15 & Translation norm/20; angle-summary norm/180\\
16--17 & Target camera ID/10 (or $-1$); mirror flag\\\bottomrule
\end{tabularx}
\end{table}

\textbf{Normalization preserves the checkpoint's coordinate contract.}
Signed ratios are clipped to $[-1,1]$; magnitudes and known camera IDs
are clipped to $[0,1]$. For captured camera transfer, the builder uses
$(T^s_0)^{-1}T^q_0$. With relative rotation $R=(r_{ab})$ and zero-based
indices, yaw, pitch, and roll are
$\operatorname{atan2}(r_{02},r_{22})$,
$\operatorname{atan2}(-r_{12},\sqrt{r_{02}^2+r_{22}^2})$, and
$\operatorname{atan2}(r_{10},r_{11})$. Mirroring negates yaw, roll, and
lateral translation. The norm of these angle summaries is an encoded
feature; per-frame ray geometry supplies the actual camera orientation.

\subsection{Optimization}
\textbf{Adaptation concentrates learning on the conditioning interface
and low-rank updates.} The pretrained Wan2.1-VACE-14B
\cite{wan2025,jiang2025vace} DiT/VACE weights,
VAE, and text encoder are frozen. We train the added modules and LoRA on
attention query/key/value/output and both FFN projections, at ranks 64/32
for DiT/VACE. LoRA alpha equals rank. VACE warmup uses learning rate
$5\times10^{-5}$; joint adaptation uses $2\times10^{-5}$. Training uses
32 B200 GPUs for approximately five days. The VACE residual scale is .85
in training and 1.0 in the component and single-scene edit runs. Joint checkpoints provide
matched LoRA, trajectory, geometry, and source-video parameters.

\textbf{The loss weights actors and missing render regions without
double-counting their overlap.} Masks are packed in $8\times8$ blocks,
temporally resampled with nearest-exact interpolation, and averaged over
packed channels. Hole and ego-ignore masks are thresholded at .3; actor
coverage remains fractional. Background weight is one, hole weight two,
and actor weight two overrides hole weighting by coverage interpolation.
Ego-ignore takes priority. The reference prefix is excluded, and the
denominator is the spatial weight sum expanded over 16 latent channels,
clamped below at one.

\textbf{Noise indices follow the pretrained shifted schedule.}
For uniformly sampled index $i\in\{0,\ldots,999\}$, set
$a_i=1-i/1000$, $\sigma_i=5a_i/(1+4a_i)$, and timestep
$s_i=1000\sigma_i$. The multiplier in the main loss is
\begin{equation}
 y_i=e^{-2(\sigma_i-1/2)^2},\qquad
 \lambda(\sigma_i)=\frac{1000(y_i-y_{\min})}
 {\sum_{j=0}^{999}(y_j-y_{\min})},
\end{equation}
where $y_{\min}$ is the minimum over schedule indices.

\section{Benchmark Protocol and Extended Results}
\label{sec:benchmark-protocol}

The main paper pairs matched baseline comparisons with synchronized
eight-heading observations. Here we detail the evaluation protocol and
extend those fixed-camera observations to continuous camera trajectories.
Video 04 shows only synchronized generated views. We rotate a virtual
camera about the moving cam0 optical center in $45^\circ$ increments.
The $0^\circ$ panel synthesizes the original heading; the seven rotated
panels test extrapolation beyond the calibrated camera poses used for
training.

\textbf{The fixed evaluation set provides paired real observations across
six target cameras.} The expanded test set contains 196 synchronized
multi-camera clips, disjoint from training under the data provider's split
audit. We average camera/window scores within each physical scene before
averaging scenes equally. All paired metrics score the common first 81 frame IDs.
The input contract comprises cam0 video, calibration, 3D tracks/boxes, and
masks. Test scenes follow the provider's fixed evaluation set, not a subset
selected by the reported method's scores.

\textbf{Held-out target images do not enter test-world construction.}
DGGT and depth processing use cam0 image observations; archived Asset
Harvester vehicles are completed from cam0 crops, and human appearance and
pose are obtained from cam0 observations. Neither preprocessing nor cache
creation reads cam1--6 RGB. The source-built geometry and assets are reused
for every target, without target-image asset selection or fitting.
Target intrinsics and poses define rendering and camera conditions;
target RGB and evaluation masks are used only in scoring and visualization.
This is a single-source-RGB protocol with the declared auxiliary annotations,
not a claim that calibration and actor tracks are recovered from RGB alone.
Synchronized target RGB supervises training only on the disjoint training
split. Editing demonstrations with donor assets or altered references are
separate from this paired benchmark.

\textbf{Actor membership is source-observed; retained motion is globally tracked.}
We use boxes and tracks only for actors that appear in cam0 observations.
Objects never observed in cam0 are not instantiated merely because they
have global annotations. Boxes initialize metric extent and placement;
per-asset alignment then refines the fit to source observations, without
using target RGB. The retained tracks continue when an observed actor
leaves cam0, for example when the ego vehicle overtakes it. This preserves
the recorded actor's subsequent motion instead of assuming constant speed
or inventing an unseen turn. It is trajectory-conditioned re-observation,
not prediction of unobserved behavior from source RGB alone.

\textbf{Restricted annotation use still supplies extra information.}
Source-based membership prevents adding actors never seen by the input
camera, but the retained positions, dimensions, headings, associations,
and out-of-view motion remain additional geometric evidence. These inputs
do not specify completed surfaces or target-view appearance. Table~\ref{tab:input-roles}
separates them from held-out target-image supervision. The protocol keeps
the observed participants' event fixed; global tracks are not a universal
requirement for evaluating novel views, nor does their restriction make
the input budget identical to video-only baselines.

\begin{table}[tbp]\centering\small
\caption{\textbf{Image evidence and auxiliary geometry have distinct roles.}
Information used by RigRecast in the captured-view benchmark.}
\label{tab:input-roles}
\begin{tabularx}{\linewidth}{@{}lX@{}}\toprule
Information & Role\\\midrule
Cam0 RGB & Source appearance, reconstruction, asset crops, references, and full-video context.\\
Calibration & Source projection and metric alignment; target intrinsics and poses define each query.\\
Restricted global boxes/tracks & Initialization and continued motion of source-observed actors only; no annotation-only actors.\\
Source masks & Foreground separation and reference availability.\\
Test target RGB & Scoring and visualization only, not test-world construction or inference.\\
Target ego masks & Identical scoring-region exclusion for output and captured GT.\\\bottomrule
\end{tabularx}
\end{table}

\textbf{Method-specific interfaces are preserved in the reproduced
comparisons.} GEN3C, NeoVerse, AnyView, and TrajectoryCrafter use their
video/geometry and camera-control routes. AnyView uses three 41-frame
segments and TrajectoryCrafter three overlapping 49-frame contexts,
resolving overlaps before scoring. NeoVerse retains source intrinsics.
ReCamMaster uses the cam1 source-FOV adaptation and native source FOV for
the remaining targets; ReCamDriving uses the source-intrinsic real-rig
extension. These interface adaptations are retained in the main-table
footnotes rather than pooled into an identical-input claim.
The comparison therefore measures complete systems on matched captured
targets, with their respective input interfaces and training histories.
It does not isolate architectural gains from supplied annotations or
task-specific adaptation. The component study instead holds the
additional training budget and scene subset fixed within our system;
variant-specific settings are detailed below.

\textbf{Independent segmentation measures traffic layout symmetrically.}
The same YOLOv8x/SAM2 \cite{ultralytics2023yolov8,ravi2024sam2} processing extracts vehicle, pedestrian, and
two-wheeler unions from output and GT. Target ego masks expanded upward
by ten pixels exclude identical image regions on both sides. Empty
unions and segmentation failures follow the same validity rule.
FID/FVD use matched frames, preprocessing, and ego filling. The all-rig
FID pools frame features per distribution; FVD pools video-clip features.
Both are computed from pooled matched features, not averages of separate
front/rear FID or FVD scores.

\begin{table}[tbp]\centering
\caption{\textbf{All captured targets: cam1--6.} The expanded 196-clip
test set, with the common first 81 frames. Input/FOV adaptations follow
the main table; all listed methods have completed all six targets.}
\begingroup\small\setlength{\tabcolsep}{4pt}\begin{tabularx}{\linewidth}{@{}Xrrrrrr@{}}\toprule
\metrichead\\\midrule
GEN3C & 0.0520 & 11.54 & 0.2622 & 0.7795 & 188.67 & 2329.73\\
NeoVerse & 0.2299 & 10.90 & 0.2850 & 0.7401 & 164.24 & 3332.10\\
AnyView & 0.1157 & 12.57 & 0.3879 & 0.7221 & 98.14 & 960.36\\
TrajectoryCrafter & 0.1412 & 9.61 & 0.2000 & 0.7734 & 184.81 & 4302.33\\
ReCamMaster & 0.1069 & 12.55 & 0.3982 & 0.7540 & 136.06 & 1099.11\\
ReCamDriving & 0.1655 & 11.66 & 0.3301 & 0.6958 & 104.74 & 1072.28\\
\method{} & \textbf{0.5585} & \textbf{14.42} & \textbf{0.4362} & \textbf{0.5470} & \textbf{39.65} & \textbf{496.89}\\
\bottomrule\end{tabularx}
\endgroup\label{tab:full-rig}
\end{table}
\begin{table}[tbp]\centering
\caption{\textbf{Lateral and rear targets: cam2/5/6.} The same expanded
test set and common first-81-frame protocol. Best values are bold.}
\begingroup\small\setlength{\tabcolsep}{4pt}\begin{tabularx}{\linewidth}{@{}Xrrrrrr@{}}\toprule
\metrichead\\\midrule
GEN3C & 0.0000 & 11.30 & 0.2740 & 0.8241 & 287.87 & 4104.49\\
NeoVerse (source K) & 0.1471 & 9.59 & 0.2568 & 0.7943 & 253.44 & 5713.68\\
AnyView ($3\times41$) & 0.0811 & 12.43 & 0.3823 & 0.7277 & 115.29 & 1295.81\\
TrajectoryCrafter ($3\times49$) & 0.1169 & 8.71 & 0.1784 & 0.7968 & 259.68 & 7374.09\\
ReCamMaster$^{\dagger}$ & 0.0811 & 12.23 & 0.3855 & 0.7485 & 153.15 & 1443.70\\
ReCamDriving$^{\ddagger}$ & 0.1350 & 11.40 & 0.3200 & 0.7052 & 119.83 & 1213.37\\
\method{} & \textbf{0.4845} & \textbf{13.45} & \textbf{0.4080} & \textbf{0.5983} & \textbf{66.10} & \textbf{850.75}\\
\bottomrule\end{tabularx}
\par\vspace{3pt}{\scriptsize\raggedright
$\dagger$ ReCamMaster: source-FOV adaptation for cam1; native source FOV for other cameras.
$\ddagger$ ReCamDriving: real-rig extension with source-$K$ geometry.\par}
\endgroup\label{tab:rear}
\end{table}
\textbf{Calibrated optical axes define the viewpoint groups.}
For camera rotations $R^s_t,R^q_t$ in a common frame and
$e_z=(0,0,1)^\top$, we compute
$\alpha_{q,t}=\arccos(\operatorname{clip}((R^s_te_z)^\top R^q_te_z,-1,1))$.
Ranges on evaluated frames are $0.37$--$1.39^\circ$ for cam1,
$68.89$--$71.03^\circ$ for cam3/4, $139.38$--$141.39^\circ$ for cam2/5,
and $177.24$--$178.95^\circ$ for cam6. Mounting translations and focal
lengths retain their real calibrated values.

\textbf{Physical-scene resampling quantifies variation across the test
set.} We bootstrap physical scene IDs with replacement for 10,000 resamples
using seed 20261115. For cam1/3/4, the IoU gain over NeoVerse is .3197
with percentile-95\% interval $[.2655,.3704]$; the PSNR gain over
ReCamMaster is 2.518\,dB with interval $[1.882,3.115]$. Across six
targets, the IoU gain is .3286 with interval $[.2916,.3641]$.
These intervals summarize scene variation at fixed model and generation seeds.

\subsection{Continuous Camera Controls}
\label{sec:continuous-camera-controls}
\begin{figure}[tbp]\centering
\includegraphics[width=\textwidth]{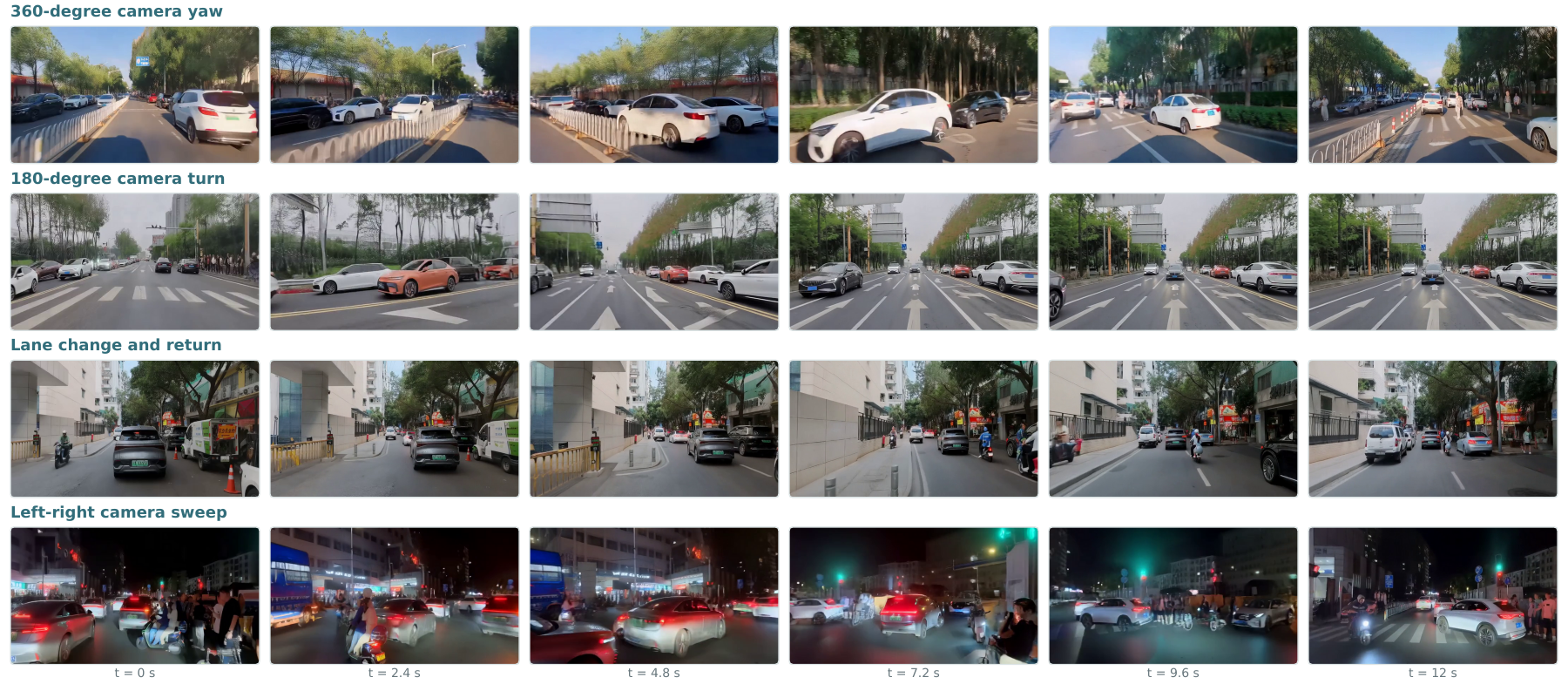}
\caption{\textbf{Four scenes, four continuous camera controls.}
Rows show a $360^\circ$ yaw sweep (daylight), a $180^\circ$ turn
(open intersection), a lane change and return (urban road), and a
left--right sweep (night). Columns sample offsets 0, 24, 48, 72, 96,
and 120 of each 121-frame video, corresponding to 0--12\,s at 10\,Hz.
Only the observation changes; actor trajectories remain fixed.
Video S03/00 shows all four trajectories alongside their recorded inputs.}
\label{fig:motion}
\end{figure}

\textbf{Continuous camera trajectories re-observe the same traffic event.}
Figure~\ref{fig:motion} samples four 121-frame sequences: a full yaw
rotation, a half turn, a lane change and return, and a left--right sweep.
Within each example, only the observation trajectory changes; actor
geometry and world-space motion remain fixed.
Video S03/00 pairs each recorded cam0 input (top) with its generated
trajectory (bottom), in the same order as the figure rows.
Videos S03/05 and S03/06 show a separate maneuver: the rig turns left
to $90^\circ$ while the ego origin continues along its original path.
S03/05 juxtaposes the recorded front view, the target-camera render, and its refined
output. S03/06 shows all six rigidly mounted views at matched times;
the inset in each tile indicates its orientation.

\section{Component Evidence}
\subsection{Dynamic Assets}
\begin{figure}[tbp]\centering
\includegraphics[width=.95\textwidth]{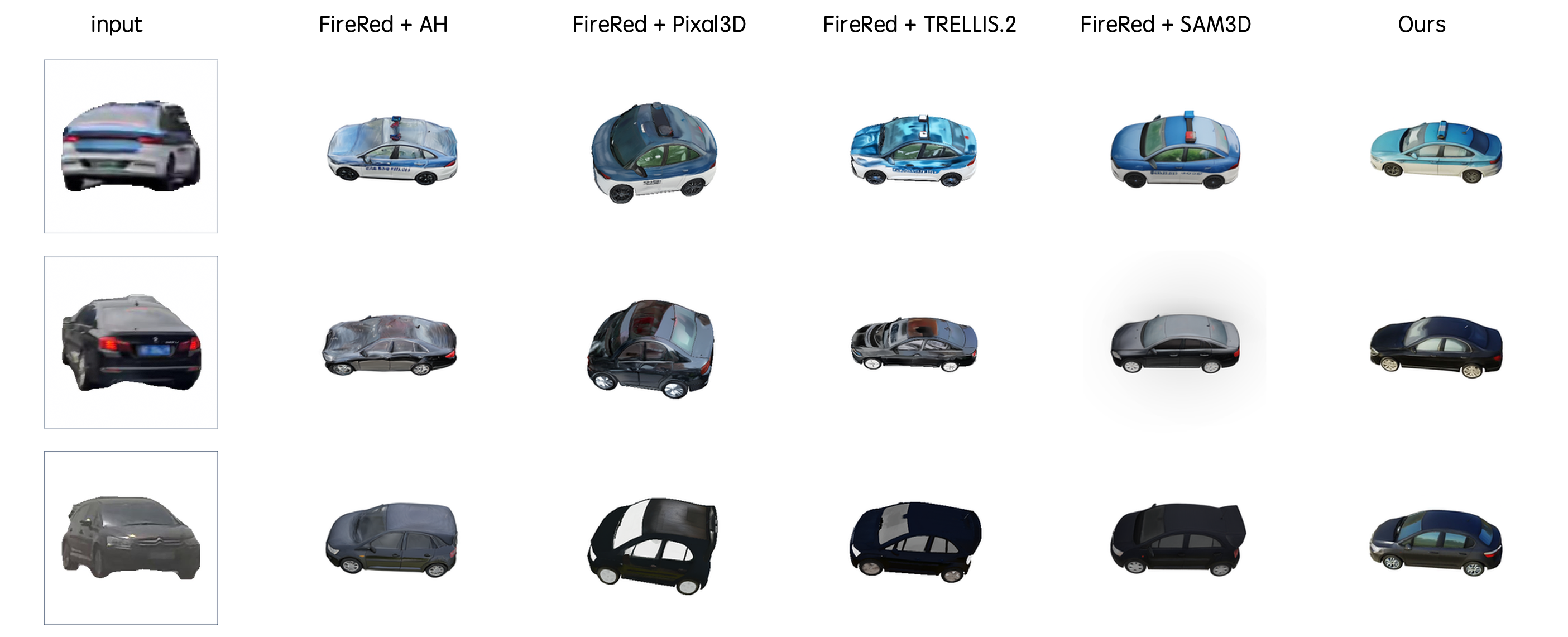}
\caption{\textbf{Directional vehicle completion.} Component examples
compare reconstructed body shape beyond the observed crop.}
\label{fig:vehicle}
\end{figure}

\begin{figure}[tbp]\centering
\includegraphics[width=.92\textwidth]{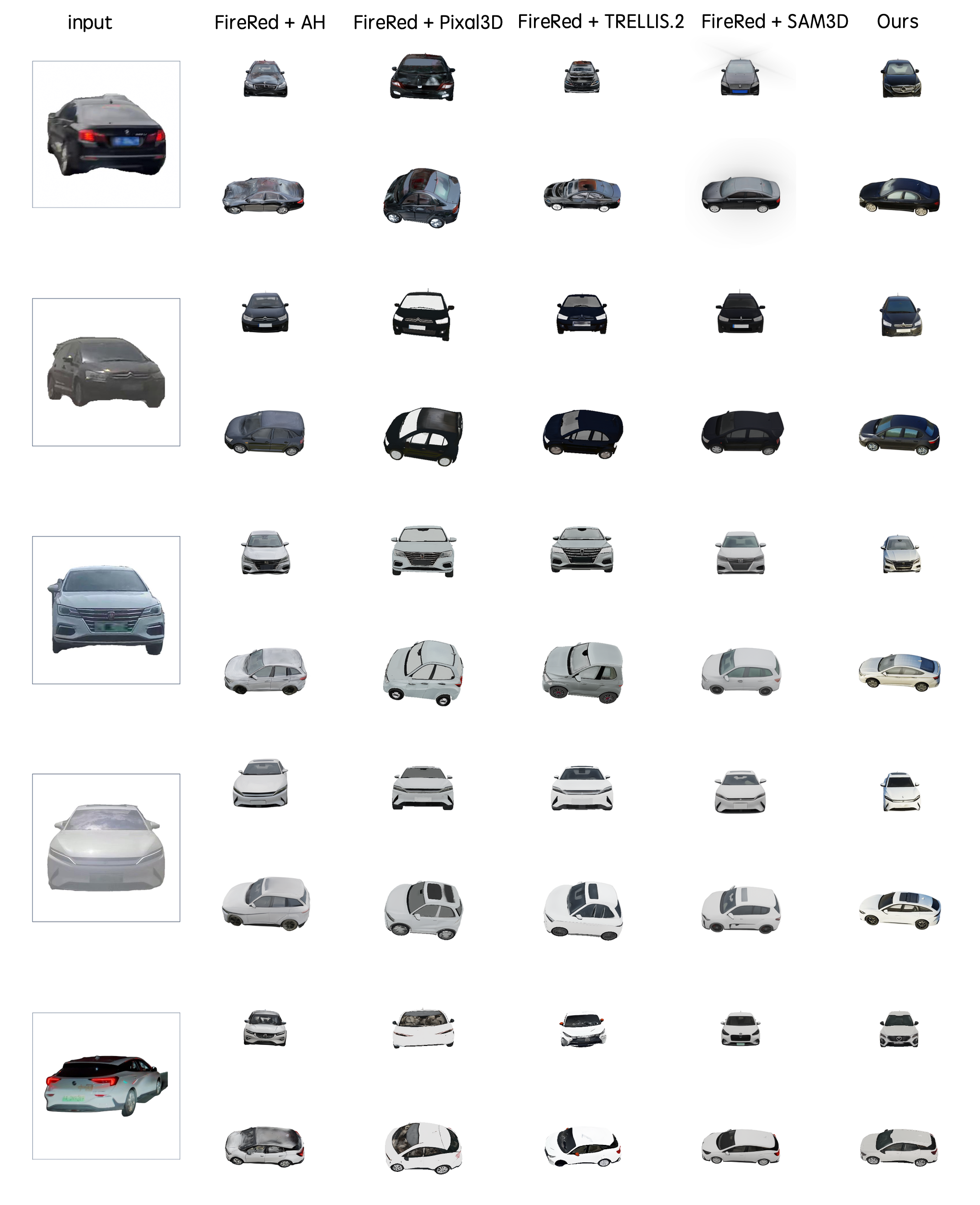}
\caption{\textbf{Additional vehicle-component examples.} Complementary
views expose body extent and surfaces missing from a single observation.}
\label{fig:vehicle-more}
\end{figure}

\begin{figure}[tbp]\centering
\includegraphics[width=.95\textwidth]{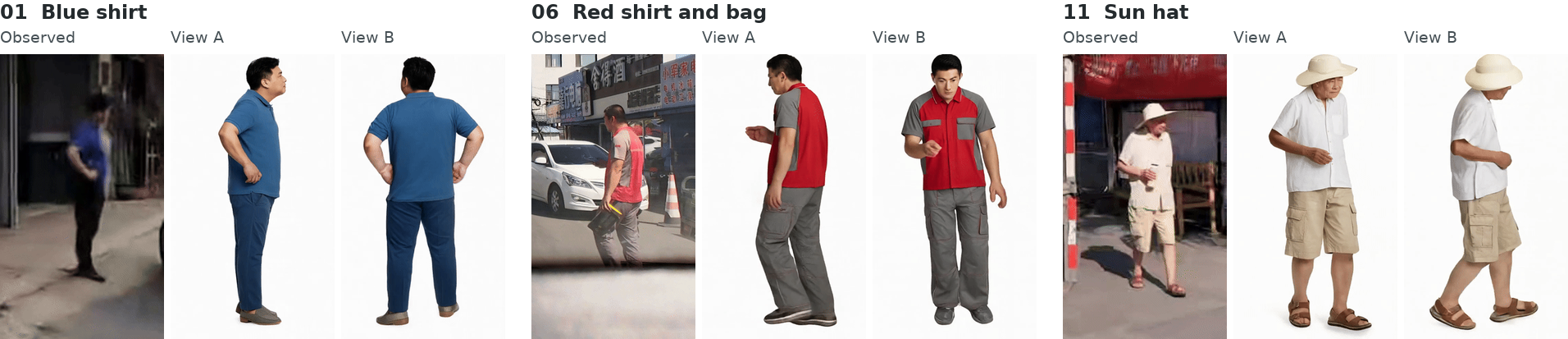}
\caption{\textbf{Articulated identities across views.} Three pedestrian
identities are shown in observed and generated views of the same posed body.}
\label{fig:human}
\end{figure}

\textbf{Completion supplies reusable bodies rather than view-specific
foreground fragments.} The directional vehicle branch cleans a crop,
constructs complementary views, reconstructs a mesh, and fits canonical
Gaussians. FireRed \cite{firered11Model} prepares crops; the Qwen-Image-Edit
multi-angle adapter \cite{qwenMultiangleModel} supplies complementary views
to adapted ReconViaGen \cite{chang2025reconviagen}. The TRELLIS.2 renderer
\cite{xiang2025trellis2} produces 160 mesh views for Gaussian fitting.
The benchmark's archived Asset Harvester cache and this
directional route use the same asset-placement interface. Pedestrian
modeling combines reference appearance with an articulated body sequence
and metric world placement. Figures~\ref{fig:vehicle}--\ref{fig:human}
show the two asset types and their different deformation requirements.
Vehicle comparisons include Asset Harvester \cite{cao2026assetharvester},
Pixal3D \cite{li2026pixal3d}, TRELLIS.2 \cite{xiang2025trellis2}, and
SAM 3D \cite{sam3d2025}.


\subsection{Static Geometry}
\textbf{The covariance intervention changes render support without moving
the scene or its actors.} Both arms retain the
same static centers, colors, opacities, dynamic assets, and original tracks.
Each static axis scale is floored at $10^{-8}$ before the geometric mean;
the spherical scale multiplier is one. A 12,794-point numerical check
preserves ellipsoid volumes within relative error $5\times10^{-6}$.

\textbf{The static covariance comparison uses the same 20 stratified test
clips as the main component study.}
Both arms receive 500 adaptation steps on the 98-clip training subset.
Evaluation uses seed 42, 30 steps, CFG 5, scale 1, and identical
source references, camera arrays, and trajectories within each clip.
Cam1/3/4 generate 121-frame sequences and score the common first 81 frames.
Across the 20 clips, mean IoU rises from .4046 to .6156, PSNR from 11.19
to 14.86\,dB, and SSIM from .3032 to .4376; LPIPS improves from .6754
to .5209. Hole coverage follows the rendered geometry, so the covariance
intervention changes both visible support and its associated hole mask.

\begin{table}[tbp]\centering\small
\caption{\textbf{Static covariance intervention over 20 test clips.}
Same test subset and 500-step adaptation budget as the main component
study; dynamic assets retained, with the first 81 frames of cam1/3/4 scored.}
\begingroup\small\setlength{\tabcolsep}{4pt}\begin{tabularx}{\linewidth}{@{}Xrrrr@{}}\toprule
Static covariance & IoU $\uparrow$ & PSNR $\uparrow$ & SSIM $\uparrow$ & LPIPS $\downarrow$\\\midrule
Original anisotropic & 0.4046 & 11.19 & 0.3032 & 0.6754\\
Equal-volume isotropic & \textbf{0.6156} & \textbf{14.86} & \textbf{0.4376} & \textbf{0.5209}\\
\bottomrule\end{tabularx}
\endgroup\label{tab:covariance}
\end{table}

\textbf{The component study compares adapted variants on a separate test subset.}
Full CCAR, LoRA-only VACE, and the five component interventions initialize
from step8000 and receive 500 additional updates at learning rate $10^{-4}$.
The 98 training clips follow the evaluation subset's stratified sampling
procedure while excluding all test scenes. Evaluation uses 20 held-out
clips, with Full scores .6156/14.86/.4376/.5209. Native VACE is the
unadapted pretrained reference and receives completed renders and four
source references. LoRA-only VACE omits the added source-video, dense-ray,
and trajectory modules but retains completed renders.

The w/o actor completion variant uses raw DGGT foregrounds with actor
loss weight one; all other adapted variants use actor loss weight two.
The native model's higher IoU with completed renders concerns layout;
its LPIPS is worse, so it does not dominate this variant on all metrics.
The adapted study remains separate from the step5250 paired benchmark
and step8000 demonstrations. Section~\ref{sec:additional-evaluation}
details the observed failure modes.

\section{World Editing Protocols and Examples}
\label{sec:edit-protocol}
\textbf{The world edit and the generated observation are recorded
separately.} Every request retains the edited asset, placement sequence,
camera parameters, render controls, and refined videos. Multiple actor
requests are applied to one world snapshot. Per-camera renders resolve
visibility and occlusion from that shared geometry; CCAR then refines
each camera video independently.

\subsection{Asset and Motion Controls}
\textbf{Paired examples separate changes in identity from changes in
motion.} Daylight cam4 substitutions replace a white sedan with a red
sedan or an SUV, and replace two tracks in one world. These use step8000,
frames 121--241, seed 42, 30 steps, CFG 5, and scale 1 at $704\times480$.
The original cam0 video and all four references remain visible, and box
dimensions and tracked motion are preserved for replacement requests.

\begin{figure}[tbp]\centering
\includegraphics[width=.90\textwidth]{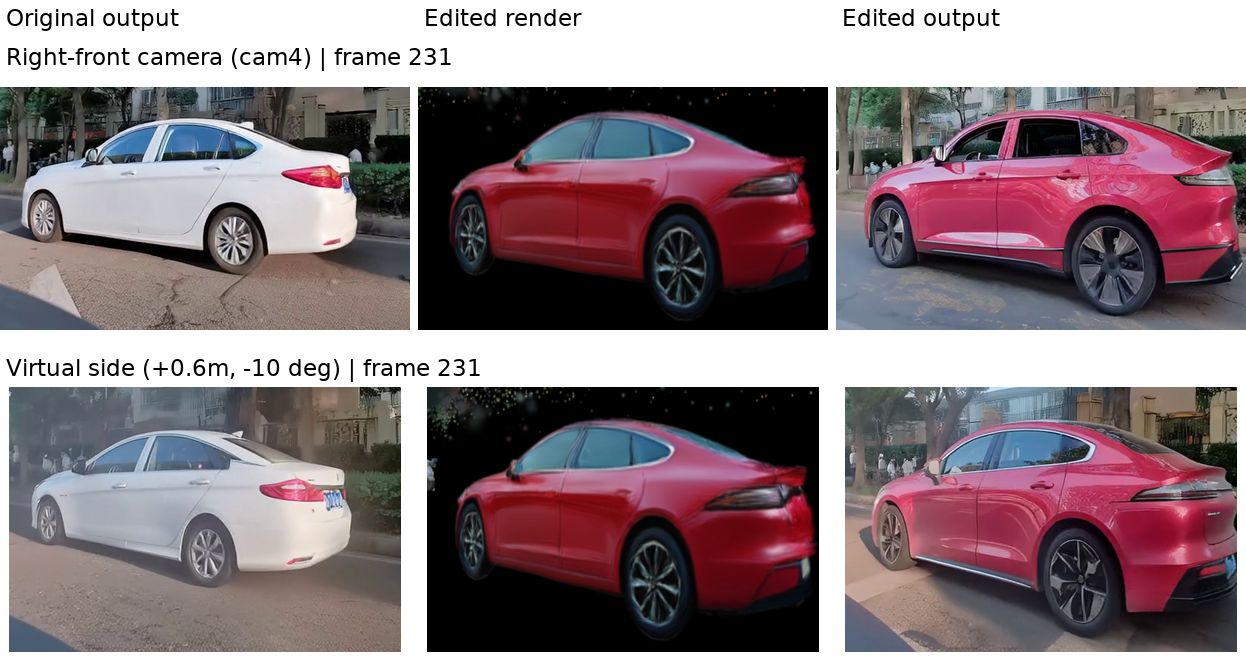}
\caption{\textbf{One asset substitution under two observations.}
Captured cam4 and a virtual side camera observe the same white-to-red
replacement at frame 231. Columns: original output, edited render, edited
output. The virtual camera shifts .6\,m along cam4's local $X$ and rotates
$-10^\circ$ about its local $Y$.}
\label{fig:shared-asset}
\end{figure}
\begin{figure}[tbp]\centering
\includegraphics[width=.93\textwidth]{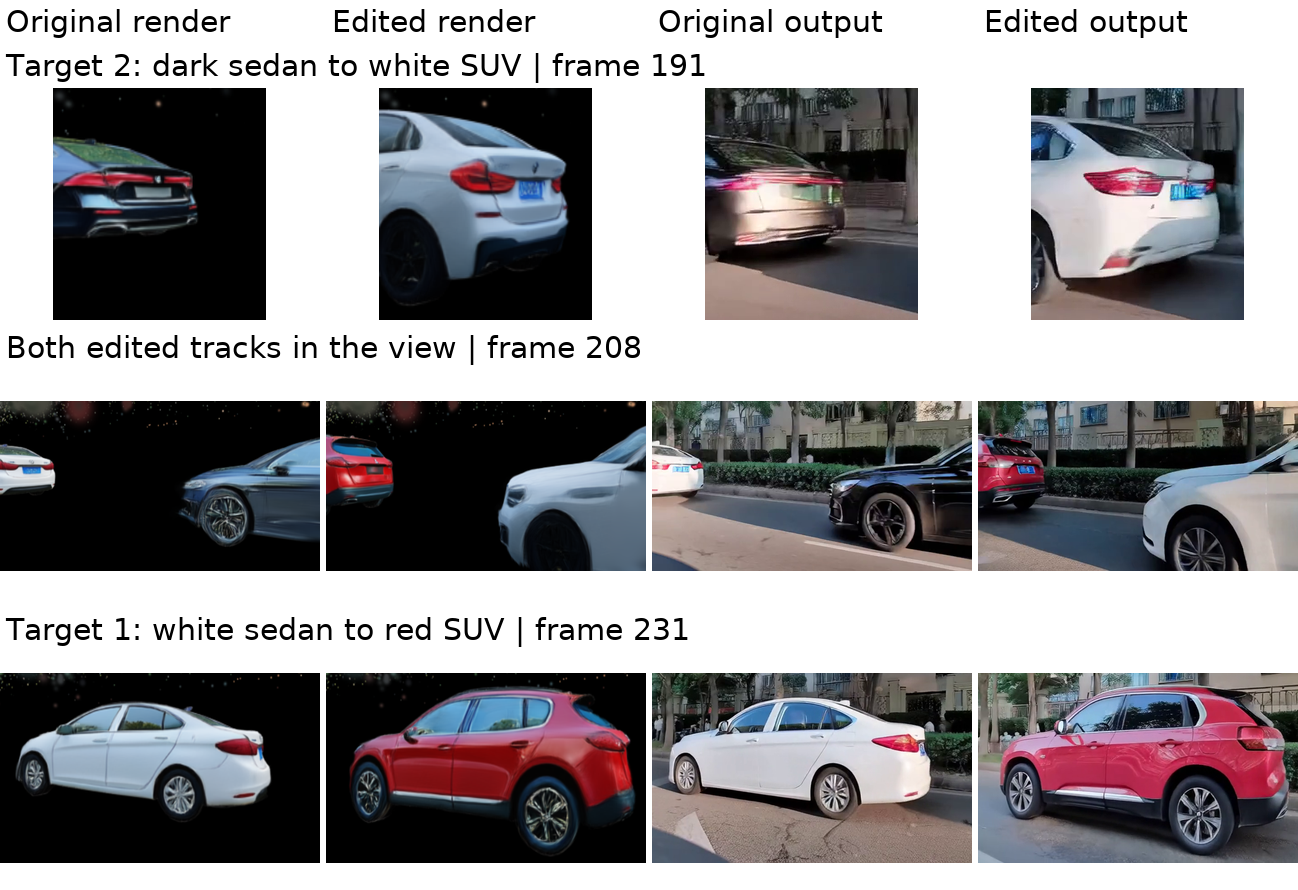}
\caption{\textbf{Two substitutions in one dynamic world.} A dark sedan
becomes a white SUV and a white sedan becomes a red SUV. Rows show frames
191, 208, and 231; columns show original render, edited render, original
output, and edited output. Both edited targets are partly visible at 208.}
\label{fig:two-actors}
\end{figure}

\textbf{A shared placement change propagates to multiple camera controls.}
The paired motion example retains the original white-sedan asset and adds
a smooth offset of 3\,m forward and 1\,m left in its heading frame at
frame 185. The offset grows through frame 225 and then stays relative to
the original track; heading follows the curved path. Captured cam4 and
the virtual side camera observe the same change.

\subsection{Six-Camera Overtaking Demonstrations}
\textbf{One inserted actor traverses several camera fields as a single
edited event.} We insert the red asset on the ego vehicle's right,
accelerate it forward, merge it left, and brake while the ego vehicle
continues. The red-vehicle example uses frames 121--241 of daylight scene
36a and removes eight path-conflicting tracks as part of its compound edit.
The yellow-vehicle example uses frames 0--120 of intersection scene c4e9.
In each example the rig and elevated views follow the same inserted
vehicle and its prescribed world trajectory. The rig uses
cam0/2/3/4/5/6; telephoto cam1 is excluded. The elevated observer is raised 8\,m with a
$45^\circ$ downward pitch, and a white vehicle marks the ego pose.

\textbf{The displayed demos include reference interventions beyond the
standard masked-cam0 protocol.} The red cam0 run
composites the edited render into original references at offsets
0/40/80/120; residual removed-object regions are marked unknown. The yellow
cam0 run uses four unmodified original references at the same offsets.
Other rig cameras use original references, and the complete cam0 video
remains unchanged. Runs use seed 42, 30 steps, CFG 5, and control scale 1.
The teaser orders cameras as right-rear, right-front, front, left-front,
left-rear, and rear. Red timestamps are 1.5/2.3/4.4/6.5/8.0/11.0\,s;
yellow timestamps are 0.8/1.0/2.2/3.0/4.2/7.2\,s. Each column pairs rig
and aerial observations at one time; columns follow the event chronologically.

\textbf{The visual comparison follows the inserted actor.}
Each video in folder 01 pairs the recorded input (upper left) with the
elevated observation (lower left) and six synchronized rig views (right).
The yellow elevated video includes revised surrounding vehicles. The red
rear and elevated observations include an auxiliary forward-oriented
vehicle behind the ego vehicle. These editing demonstrations are separate from the captured-view
benchmark, where annotation-derived actors must have source observations.

\subsection{Actor Edits with Spherical Static Context}
\textbf{Two interventions share a reconstructed intersection but use
different observation cameras.} Figure~\ref{fig:edit-pair} uses scene33,
frames 0--120. In the joint insertion, the red vehicle enters from the
right and merges while the yellow vehicle advances ahead. The virtual
observer advances with the ego motion; its camera is raised by 3\,m,
turned right by $25^\circ$, and pitched down by $15^\circ$.
The fleet experiment replaces
24 CAR-class assets with four donor vehicle types from the daylight scene,
preserving their original tracks, dimensions, and headings. Its camera
is shifted forward by 4\,m, raised by 1.5\,m, turned left by $20^\circ$,
and pitched down by $8^\circ$.

\textbf{Rendering and generation use matched geometry conditions.}
Both experiments retain spherical static context and original human and
two-wheeler tracks. Camera rays and trajectory conditions use the rendered
camera. CCAR uses step8000, seed 42, 30 steps, CFG 5, control scale 1.5,
and $704\times480$ resolution. Both virtual views retain four complete
cam0 references at 0/40/80/120 and the complete source video. Figure panels
use one common vertical crop; full-frame videos are retained. Boxes are
post-render annotations from edited-actor alpha support, not model inputs.
These qualitative experiments are separate from the ten-scene scoring pilot.
The synchronized video in folder 05 shows the recorded cam0 input on the
left. Three columns compare a single insertion, fleet replacement, and
joint insertion of red and yellow vehicles viewed from a camera raised
by 3\,m. In each column, colored corners identify edited vehicles in
the geometry render (top); the refined output (bottom) is unmarked.

\subsection{Agent-Assisted Ten-Scene Evaluation}
\label{sec:agent-edit-assessment}
\textbf{A Qwen3-VL-32B agent assesses whether generated observations
follow the requested world edits.} The B200 evaluation covers ten scenes,
84 requests, and 212 eligible request--view pairs. Each request specifies
the target actor and the intended appearance, motion, or visibility
change. The evaluator compares these requirements with the render controls
and generated video, checking the edited target, the requested attributes
or action, event order, and temporal persistence. The model is from the
Qwen3-VL family \cite{qwen3vlRepository}; a small subset of judgments is
manually adjudicated before aggregation.
Method identity is hidden from the evaluator. Videos are sampled at every
second frame, and five cases are manually reviewed.

\textbf{Four labels distinguish adherence from uncertainty.}
Success (S) means that the requested change is realized and maintained in
the eligible observation. Partial (P) means that a substantive part is
realized but an attribute, motion phase, or persistence requirement is
not met. Failure (F) means that the requested change is not realized or
the wrong target or action is produced. Unjudgeable (U) denotes
insufficient visual evidence. The evaluation follows the requested
intervention: an intentionally unusual pose or overlap is not a failure
merely because it is physically implausible. Partial outcomes remain
separate from successes, and uncertainty is not treated as a pass.

\textbf{Adherence is defined by the requested operation, not generic
video realism.} Speed changes require the intended acceleration or braking
relative to the ego motion; lateral shifts require displacement to the
requested side, not merely a heading change. Asset edits require the
specified target, appearance or type, and continued presence when visible.
Flicker on/off tests the requested presence and absence intervals. Extreme
poses test the specified orientation or lift, and overlap tests whether
the requested actors and their spatial relation persist. Completing only
part of an action is P rather than F; a clearly absent expected actor is
failure evidence rather than automatically U. Fine identity attributes
matter only when requested and visually observable. An instance IoU can
support layout assessment, but cannot establish color, type, or event order.

\begin{table}[tbp]\centering
\caption{\textbf{Standard edits transfer more reliably than stress edits.}
S/P/F/U: success/partial/failure/unjudgeable. Strict SR is S/(S+P+F).
Joint requires success in every eligible view; U is not passed.}
\label{tab:edit-adherence}
\begingroup\small\setlength{\tabcolsep}{4pt}\begin{tabularx}{\linewidth}{@{}Xrrrr@{}}\toprule
Operation & Req. & S/P/F/U & Strict (\%) & Joint\\\midrule
Speed changes & 15 & 29/1/0/0 & 96.7 & 14/15\\
Lateral shift & 8 & 16/0/0/0 & 100.0 & 8/8\\
Asset edits & 19 & 49/3/0/0 & 94.2 & 16/19\\
\textbf{Standard} & \textbf{42} & \textbf{94/4/0/0} & \textbf{95.9} & \textbf{38/42}\\
\midrule
Flicker on/off & 14 & 1/31/8/0 & 2.5 & 0/14\\
Extreme poses & 20 & 21/20/18/1 & 35.6 & 2/20\\
Overlap & 8 & 0/8/5/1 & 0.0 & 0/8\\
\textbf{Challenging} & \textbf{42} & \textbf{22/59/31/2} & \textbf{19.6} & \textbf{2/42}\\
\midrule
\textbf{Overall} & \textbf{84} & \textbf{116/63/31/2} & \textbf{55.2} & \textbf{40/84}\\
\bottomrule\end{tabularx}
\endgroup
\end{table}

\textbf{Table~\ref{tab:edit-adherence} pools view-level judgments and separately measures
joint execution.} For each operation, strict success is
$100S/(S+P+F)$. Standard edits comprise speed changes, lateral shift,
and asset edits; challenging edits comprise flicker on/off, extreme
poses, and overlap. Subtotals pool counts rather than average row
percentages. With $h_{iq}$ the label for request $i$ in eligible camera
$q$, the joint rate is
\begin{equation}
 \mathrm{JointSR}=\frac{100}{84}\sum_{i=1}^{84}
 \mathbf1[\,h_{iq}=\mathrm{S}\ \text{for all }q\in\mathcal Q_i\,],
\end{equation}
where $\mathcal Q_i$ contains that request's eligible views. The resulting
40/84 (47.6\%) joint rate retains requests containing U in the denominator.
Of 212 eligible pairs, 210 are judgeable (99.1\% coverage); strict success
is 116/210 (55.2\%), or 116/212 (54.7\%) when all eligible pairs are
retained. Joint measures execution across views, rather than a separate
pixel-level identity-consistency metric.

\subsection{Single-Scene Human Ratings and Protected Regions}
\label{sec:single-scene-edit-study}
\textbf{An earlier human-rated experiment provides a separate locality
study.} Its twelve requests and manual labels below are not the
ten-scene agent-assisted counts in the main table.
\textbf{A fixed-appearance study separates requested edits from changes
elsewhere in the output.} Twelve requests in daylight 36a use cam0,
right-front cam4, and a virtual camera raised 8\,m and pitched down
$45^\circ$. All 36 runs use step8000 and the same generation settings.
The control world contains a red sedan. Four source-view references
composite this red asset into the observed crop as known content, and
remain fixed across requests; the original full source video is unchanged.
Conditions render completed vehicles and people on an unknown black
background, without static or original dynamic scene splats.

\textbf{Requests cover identity, motion, and stress interventions.}
They include acceleration, braking, lateral motion, substitution, insertion,
removal, flicker-on appearance and flicker-off disappearance, $180^\circ$ actor yaw,
$90^\circ$ roll, a 3\,m lift, and interpenetration. A frozen projected-hull
visibility test admits 33 of 36 request--view pairs; three lead-car motion
requests (acceleration, braking, and lateral shift) are outside cam4's
evaluable field. These exclusions precede human scoring, leaving
12 cam0, nine cam4, and twelve aerial pairs.

\textbf{One author scored all eligible pairs against their requested
rendered changes.} The labels are success (S), partial (P), failure (F),
and unjudgeable (U); all 33 eligible render controls were judged valid.
Table~\ref{tab:request-scores} reports every request. For each operation
group, $\mathrm{SR}=100S/(S+P+F)$; the single U label is excluded from
this denominator. Overall, 17/32 judgeable pairs succeed (53.1\%);
the verified-success fraction over all 33 eligible pairs is 51.5\%.
Partial responses remain separate from successes. These are descriptive
ratings from one rater on one scene, not a multi-scene success estimate.

\textbf{Grouped scores distinguish standard edits from challenging
interventions.} Speed changes, lateral shift, and asset edits form the
standard group: 14 successes and one partial response across 15
eligible pairs, giving 14/15 (93.3\%). Flicker on/off, extreme poses,
and overlap form the challenging group: three successes, nine partial
responses, five failures, and one unjudgeable pair, giving 3/17
(17.6\%) over judgeable pairs. Each subtotal pools the underlying
judgments, rather than averaging category-level percentages. All six
categories remain in the overall score.

\begin{table}[tbp]\centering
\caption{\textbf{Per-request human judgments.} C0: source view; C4:
right-front; Air: virtual aerial. A dash is a predeclared visibility
exclusion. Joint requires S in every eligible view; N means not passed.}
\begingroup\small\setlength{\tabcolsep}{4pt}\begin{tabularx}{\linewidth}{@{}Xcccc@{}}\toprule
Operation & Cam0 & Cam4 & Aerial & Joint\\\midrule
Accelerate & S & -- & S & S\\
Brake to stop & S & -- & S & S\\
Lateral shift & S & -- & S & S\\
Substitute asset & S & S & S & S\\
Insert vehicle & S & S & S & S\\
Remove vehicle & P & S & S & N\\
Flicker-on appearance & P & P & F & N\\
Flicker-off disappearance & P & P & P & N\\
Yaw $180^\circ$ & S & S & P & N\\
Roll $90^\circ$ & P & F & P & N\\
Lift 3 m & F & F & S & N\\
Interpenetration & P & F & U & U\\
\bottomrule\end{tabularx}
\endgroup\label{tab:request-scores}
\end{table}

\textbf{Joint success requires the same requested control to pass in
every eligible observation.} With $h_{iq}$ the S/P/F/U judgment for
request $i$ and camera $q$, and $\mathcal Q_i$ its eligible cameras,
\begin{equation}
 \mathrm{JointSR}=\frac{100}{12}\sum_{i=1}^{12}
 \mathbf1[\,h_{iq}=\mathrm{S}\ \text{for all }q\in\mathcal Q_i\,].
\end{equation}
Five requests pass, six do not, and one contains an unjudgeable view;
retaining all requests gives 5/12 (41.7\%). The standard group passes
5/6 requests (83.3\%); the challenging group passes 0/6, retaining its
unjudgeable request in the denominator. This tests cross-view
execution of a request, not exact appearance identity. The separate
fine-identity labels for asset replacement are partial at cam0 and
successful at cam4 and the aerial camera, even though the requested
replacement operation is rated successful in all three.

\textbf{Protected-region LPIPS measures collateral appearance change.}
For each frame we dilate the union of the original and edited target
3D projections by radius $d\in\{32,64\}$ pixels. This region is filled
with RGB 127 in both refined images. Spatial AlexNet LPIPS is averaged
over the remaining pixels and post-onset frames. The overall statistic
first averages eligible views within a request, then the twelve requests.
Table~\ref{tab:preservation} reports the resulting $P_d$ values.
Protected pixels can contain unmodified actors as well as background;
the exclusion is the before/after footprint union, not only the edited
target's final location.

\begin{table}[tbp]\centering
\caption{\textbf{Appearance change outside edited footprints.}
Lower protected-region LPIPS means less collateral change.}
\begingroup\small\setlength{\tabcolsep}{4pt}\begin{tabularx}{\linewidth}{@{}Xrrr@{}}\toprule
Camera & Pairs & $P_{32}\downarrow$ & $P_{64}\downarrow$\\\midrule
Source cam0 & 12 & 0.1279 & 0.1263\\
Right front & 9 & 0.3937 & 0.3816\\
Virtual aerial & 12 & 0.1430 & 0.1403\\
\midrule Request-balanced & 33 & 0.1983 & 0.1939\\\bottomrule\end{tabularx}
\endgroup\label{tab:preservation}
\end{table}

\textbf{The right-front distance remains larger on a matched request set.}
On the nine requests evaluable in all three views, mean $P_{32}$ is
.1306 at cam0, .3937 at cam4, and .1538 aerially. Thus unequal view counts
alone do not explain the difference. Each control/edit pair uses the same
seed 42, so the difference is not caused by explicitly changing the random
seed. The largest cam4 distances occur for disappearance, overlap,
removal, and roll. These measurements characterize appearance changes;
they neither determine editing success nor identify a single cause for
the right-front discrepancy.

\subsection{No-Edit Resampling Control}
\textbf{A seed-only control calibrates the scale of stochastic appearance
change.} We repeat the 36a cam4 control with seed 43, leaving the control
world, rendered frames, masks, composited references, source video,
camera arrays, checkpoint, and sampling parameters unchanged from seed 42.
As in the scoring pilot, the render contains foreground assets without
static background Gaussians, leaving static surroundings to synthesis.
The control world already contains the reference red sedan; ``no edit''
means no further change to that world. Both runs contain 121 frames at
$704\times480$, with 30 steps, CFG 5, and control scale 1.

\textbf{The comparison uses exactly the same protected pixels.} For the
asset-replacement request, the control and replacement target hulls define
one per-frame exclusion union, dilated by 32 or 64 pixels. We apply this
same union and RGB-127 neutralization to both comparisons in
Table~\ref{tab:seed-control}, averaging the same 121 frames. This is one
seed pair from one camera, not an estimate of sampling variance or a
replacement for the request-averaged locality table. LPIPS differences
are not additive causal effects and are not subtracted to form a score.
In this pair, resampling produces larger protected-region distances than
the matched same-seed asset replacement.

\begin{table}[tbp]\centering
\caption{\textbf{Matched-footprint resampling check.} One cam4 seed pair
and the asset-replacement request, both relative to control seed 42.
Lower is more similar in the protected region.}
\begingroup\small\setlength{\tabcolsep}{4pt}\begin{tabularx}{\linewidth}{@{}Xrr@{}}\toprule
Comparison to control seed 42 & $P_{32}$ & $P_{64}$\\\midrule
No edit, seed 43 & 0.5392 & 0.5324\\
Asset replacement, seed 42 & 0.3024 & 0.2981\\
\bottomrule\end{tabularx}
\endgroup\label{tab:seed-control}
\end{table}

\section{Adaptation Diagnostics}
\label{sec:additional-evaluation}
Adapted component variants initialize
from step8000 at learning rate $10^{-4}$ and use 98 training clips
selected by the same stratified sampling procedure as the evaluation
subset, but with no test-scene overlap. Evaluation remains on 20 held-out
clips.
The common trained initialization makes this a study of adaptation after
component changes. It is distinct from training each architecture
independently from the pretrained backbone.

\paragraph{Adaptation does not replace geometric support.}
The w/o actor completion variant retains raw DGGT foreground geometry.
After 500 updates, its IoU is .3215 versus .6156 for Full CCAR and .4485
for Native VACE with completed renders under the respective settings
specified in the component study.

\paragraph{Plausibility and observation fidelity differ.}
LoRA-only VACE and Full CCAR both receive 500 adaptation steps on the
98-clip training subset. LoRA-only VACE lacks the source-video, dense-ray,
and trajectory-token additions. Its videos can look plausible while
deviating from the requested viewpoint, apparent camera motion, and source
appearance. Full CCAR improves IoU from .4612 to .6156 and LPIPS from
.7930 to .5209. This comparison concerns adaptation with the respective
interfaces, not an inference-only removal of trained conditions.

\paragraph{Rays and the operation token stabilize sparse-support views.}
The adapted ray-free and token-free variants reach .5763 and .5517 IoU,
respectively on cam1/3/4. Separate qualitative side-rear sequences show
incorrect viewpoints when slow ego motion limits rear observations and
leaves the render sparse. Without
the trajectory token, apparent speed changes and large unintended turns
also occur. The per-frame source/target rays remain present in this
variant. The visual behavior suggests that the operation summary helps
stabilize the intended camera motion when rendered evidence is weak,
complementing the per-frame ego-motion information still carried by rays.

\paragraph{Source-video context preserves appearance.}
Without the source-video adapter, camera conditions and four source
references remain, but illumination and texture weaken. LPIPS increases
from .5209 to .6680 and PSNR decreases from 14.86 to 13.25\,dB.
These observations distinguish appearance degradation from the prominent
viewpoint and motion errors in the two camera-conditioning ablations.

\paragraph{Static support affects apparent ground contact.}
In the covariance comparison, stretched background splats can cover
vehicles and pedestrians. Ground splats may appear above the intended
surface, hiding vehicle undersides and lower limbs although actor
placements are unchanged. Sphericalization reduces this false occlusion
and exposes more of an approximately grounded asset to the refiner.
Actor elevation stays fixed, so the observed gain concerns visibility
of the existing placement. This explains why static covariance matters
even when completed dynamic geometry is retained.

\clearpage
{\small\bibliographystyle{ieeenat_fullname}\bibliography{references}}

\begin{thebibliography}{47}
\providecommand{\natexlab}[1]{#1}
\providecommand{\url}[1]{\texttt{#1}}
\expandafter\ifx\csname urlstyle\endcsname\relax
  \providecommand{\doi}[1]{doi: #1}\else
  \providecommand{\doi}{doi: \begingroup \urlstyle{rm}\Url}\fi

\bibitem[Bai et~al.(2025)Bai, Xia, Fu, Wang, Mu, Cao, Liu, Hu, Bai, Wan, and Zhang]{bai2025recammaster}
Jianhong Bai, Menghan Xia, Xiao Fu, Xintao Wang, Lianrui Mu, Jinwen Cao, Zuozhu Liu, Haoji Hu, Xiang Bai, Pengfei Wan, and Di Zhang.
\newblock {ReCamMaster}: Camera-controlled generative rendering from a single video.
\newblock \emph{arXiv preprint arXiv:2503.11647}, 2025.

\bibitem[Cao et~al.(2026{\natexlab{a}})Cao, Ren, Zhang, Seo, Huang, Solanki, Zhang, Guo, Turki, Li, Zhu, Zhang, Gojcic, Fidler, and Yin]{cao2026assetharvester}
Tianshi Cao, Jiawei Ren, Yuxuan Zhang, Jaewoo Seo, Jiahui Huang, Shikhar Solanki, Haotian Zhang, Mingfei Guo, Haithem Turki, Muxingzi Li, Yue Zhu, Sipeng Zhang, Zan Gojcic, Sanja Fidler, and Kangxue Yin.
\newblock Asset harvester: Extracting 3d assets from autonomous driving logs for simulation.
\newblock \emph{arXiv preprint arXiv:2604.18468}, 2026{\natexlab{a}}.

\bibitem[Cao et~al.(2026{\natexlab{b}})Cao, Zhang, Voleti, Wu, B~R, Vainer, Boss, and Liu]{cao2026fourdirector}
Wei Cao, Hao Zhang, Vikram Voleti, Yuqun Wu, Mallikarjun B~R, Shimon Vainer, Mark Boss, and Yaoyao Liu.
\newblock 4director: Controlling video world models with rigid 3d geometry.
\newblock \emph{arXiv preprint arXiv:2610.02160}, 2026{\natexlab{b}}.

\bibitem[Chang et~al.(2025)Chang, Ye, Wu, Chen, Zhang, Luo, Li, Zhi, and Han]{chang2025reconviagen}
Jiahao Chang, Chongjie Ye, Yushuang Wu, Yuantao Chen, Yidan Zhang, Zhongjin Luo, Chenghong Li, Yihao Zhi, and Xiaoguang Han.
\newblock {ReconViaGen}: Towards accurate multi-view 3d object reconstruction via generation.
\newblock \emph{arXiv preprint arXiv:2510.23306}, 2025.

\bibitem[Chen et~al.(2026)Chen, Xiong, Chen, Li, Wang, Luo, Chen, Sun, Wang, Chen, Li, Zhang, Ye, and Zhao]{chen2026dggt}
Xiaoxue Chen, Ziyi Xiong, Yuantao Chen, Gen Li, Nan Wang, Hongcheng Luo, Long Chen, Haiyang Sun, Bing Wang, Guang Chen, Hongyang Li, Ya-Qin Zhang, Hangjun Ye, and Hao Zhao.
\newblock {DGGT}: Feedforward 4d reconstruction of dynamic driving scenes using unposed images.
\newblock In \emph{Proceedings of the IEEE/CVF Conference on Computer Vision and Pattern Recognition}, pages 1265--1276, 2026.

\bibitem[{FireRed Team}(2026)]{firered11Model}
{FireRed Team}.
\newblock {FireRed-Image-Edit-1.1}.
\newblock Hugging Face model card, 2026.
\newblock Model version 1.1; accessed November 17, 2026.

\bibitem[Gao et~al.(2023)Gao, Chen, Xie, Hong, Li, Yeung, and Xu]{gao2023magicdrive}
Ruiyuan Gao, Kai Chen, Enze Xie, Lanqing Hong, Zhenguo Li, Dit-Yan Yeung, and Qiang Xu.
\newblock {MagicDrive}: Street view generation with diverse 3d geometry control.
\newblock \emph{arXiv preprint arXiv:2310.02601}, 2023.

\bibitem[Han et~al.(2026)Han, Shi, Wang, Xu, Pu, Li, Zhang, Yin, and Long]{han2026enkg}
Yizhao Han, Tianxing Shi, Zhao Wang, Zifan Xu, Zhiyuan Pu, Mingxiao Li, Qian Zhang, Wei Yin, and Xiao-Xiao Long.
\newblock Entropy-guided {k-Guard} sampling for long-horizon autoregressive video generation.
\newblock \emph{arXiv preprint arXiv:2601.19488}, 2026.

\bibitem[He et~al.(2024)He, Xu, Guo, Wetzstein, Dai, Li, and Yang]{he2024cameractrl}
Hao He, Yinghao Xu, Yuwei Guo, Gordon Wetzstein, Bo Dai, Hongsheng Li, and Ceyuan Yang.
\newblock Cameractrl: Enabling camera control for text-to-video generation.
\newblock \emph{arXiv preprint arXiv:2404.02101}, 2024.

\bibitem[Heusel et~al.(2017)Heusel, Ramsauer, Unterthiner, Nessler, and Hochreiter]{heusel2017fid}
Martin Heusel, Hubert Ramsauer, Thomas Unterthiner, Bernhard Nessler, and Sepp Hochreiter.
\newblock Gans trained by a two time-scale update rule converge to a local nash equilibrium.
\newblock In \emph{Advances in Neural Information Processing Systems}, 2017.

\bibitem[Hu et~al.(2021)Hu, Shen, Wallis, Allen-Zhu, Li, Wang, Wang, and Chen]{hu2021lora}
Edward~J. Hu, Yelong Shen, Phillip Wallis, Zeyuan Allen-Zhu, Yuanzhi Li, Shean Wang, Lu Wang, and Weizhu Chen.
\newblock Lora: Low-rank adaptation of large language models.
\newblock \emph{arXiv preprint arXiv:2106.09685}, 2021.

\bibitem[Jiang et~al.(2026)Jiang, Ma, Pan, and Zhang]{jiang2026driveweaver}
Junzhe Jiang, Zipei Ma, Zijie Pan, and Li Zhang.
\newblock {DriveWeaver}: Point-conditioned video inpainting for controllable vehicle insertion in autonomous driving simulation.
\newblock In \emph{European Conference on Computer Vision}, 2026.

\bibitem[Jiang et~al.(2025)Jiang, Han, Mao, Zhang, Pan, and Liu]{jiang2025vace}
Zeyinzi Jiang, Zhen Han, Chaojie Mao, Jingfeng Zhang, Yulin Pan, and Yu Liu.
\newblock {VACE}: All-in-one video creation and editing.
\newblock \emph{arXiv preprint arXiv:2503.07598}, 2025.

\bibitem[Kerbl et~al.(2023)Kerbl, Kopanas, Leimk{\"u}hler, and Drettakis]{kerbl2023gaussians}
Bernhard Kerbl, Georgios Kopanas, Thomas Leimk{\"u}hler, and George Drettakis.
\newblock 3d gaussian splatting for real-time radiance field rendering.
\newblock \emph{ACM Transactions on Graphics}, 42\penalty0 (4), 2023.

\bibitem[Li et~al.(2026{\natexlab{a}})Li, Zhao, Chen, Hu, Guo, Zhang, Shan, and Hu]{li2026pixal3d}
Dong-Yang Li, Wang Zhao, Yuxin Chen, Wenbo Hu, Meng-Hao Guo, Fang-Lue Zhang, Ying Shan, and Shi-Min Hu.
\newblock {Pixal3D}: Pixel-aligned 3d generation from images.
\newblock \emph{arXiv preprint arXiv:2605.10922}, 2026{\natexlab{a}}.

\bibitem[Li et~al.(2026{\natexlab{b}})Li, Wang, Guo, Huang, Ding, Hu, Wang, Shen, and Tan]{li2025recamdriving}
Yaokun Li, Shuaixian Wang, Mantang Guo, Jiehui Huang, Taojun Ding, Mu Hu, Kaixuan Wang, Shaojie Shen, and Guang Tan.
\newblock {ReCamDriving}: {LiDAR}-free camera-controlled video synthesis for novel trajectories.
\newblock In \emph{European Conference on Computer Vision (ECCV)}, 2026{\natexlab{b}}.

\bibitem[Liang et~al.(2024)Liang, Yan, Chen, Zhou, Yan, Zhong, and Zou]{liang2024driveeditor}
Yiyuan Liang, Zhiying Yan, Liqun Chen, Jiahuan Zhou, Luxin Yan, Sheng Zhong, and Xu Zou.
\newblock {DriveEditor}: A unified 3d information-guided framework for controllable object editing in driving scenes.
\newblock \emph{arXiv preprint arXiv:2412.19458}, 2024.

\bibitem[Liang et~al.(2026)Liang, Yan, Zhang, Liu, Lin, Zou, Sang, and Gao]{liang2026scenedirector}
Yiyuan Liang, Zhiying Yan, Tao Zhang, Shangke Liu, Kai Lin, Xu Zou, Nong Sang, and Changxin Gao.
\newblock {SceneDirector}: Bridging explicit geometry and generative priors for unified driving scene editing.
\newblock In \emph{Proceedings of the 43rd International Conference on Machine Learning}, pages 73631--73651, 2026.

\bibitem[Lipman et~al.(2022)Lipman, Chen, Ben-Hamu, Nickel, and Le]{lipman2022flow}
Yaron Lipman, Ricky T.~Q. Chen, Heli Ben-Hamu, Maximilian Nickel, and Matt Le.
\newblock Flow matching for generative modeling.
\newblock \emph{arXiv preprint arXiv:2210.02747}, 2022.

\bibitem[Mildenhall et~al.(2020)Mildenhall, Srinivasan, Tancik, Barron, Ramamoorthi, and Ng]{mildenhall2020nerf}
Ben Mildenhall, Pratul~P. Srinivasan, Matthew Tancik, Jonathan~T. Barron, Ravi Ramamoorthi, and Ren Ng.
\newblock {NeRF}: Representing scenes as neural radiance fields for view synthesis.
\newblock In \emph{European Conference on Computer Vision}, 2020.

\bibitem[Odin(2026)]{qwenMultiangleModel}
Lovis Odin.
\newblock {Qwen-Image-Edit-2511-Multiple-Angles-LoRA}.
\newblock Hugging Face model card, published by fal, 2026.
\newblock Base model: Qwen-Image-Edit-2511; accessed November 17, 2026.

\bibitem[Qiu et~al.(n.d.)Qiu, Li, Li, Zuo, Gu, Dong, Yuan, Peng, Zhu, Han, Chen, and Dong]{lhmplusplusRepository}
Lingteng Qiu, Peihao Li, Heyuan Li, Qi Zuo, Xiaodong Gu, Yuan Dong, Weihao Yuan, Rui Peng, Siyu Zhu, Xiaoguang Han, Guanying Chen, and Zilong Dong.
\newblock {LHM++}: An efficient large human reconstruction model for pose-free images to 3d.
\newblock Official code repository, n.d.
\newblock Accessed November 17, 2026.

\bibitem[{Qwen Team}(2025)]{qwen3vlRepository}
{Qwen Team}.
\newblock {Qwen3-VL}.
\newblock Official model repository, 2025.
\newblock Accessed November 17, 2026.

\bibitem[Ravi et~al.(2024)Ravi, Gabeur, Hu, Hu, Ryali, Ma, Khedr, R{\"a}dle, Rolland, Gustafson, Mintun, Pan, Alwala, Carion, Wu, Girshick, Doll{\'a}r, and Feichtenhofer]{ravi2024sam2}
Nikhila Ravi, Valentin Gabeur, Yuan-Ting Hu, Ronghang Hu, Chaitanya Ryali, Tengyu Ma, Haitham Khedr, Roman R{\"a}dle, Chloe Rolland, Laura Gustafson, Eric Mintun, Junting Pan, Kalyan~Vasudev Alwala, Nicolas Carion, Chao-Yuan Wu, Ross Girshick, Piotr Doll{\'a}r, and Christoph Feichtenhofer.
\newblock Sam 2: Segment anything in images and videos.
\newblock \emph{arXiv preprint arXiv:2408.00714}, 2024.

\bibitem[Ren et~al.(2025)Ren, Shen, Huang, Ling, Lu, Nimier-David, M{\"u}ller, Keller, Fidler, and Gao]{ren2025gen3c}
Xuanchi Ren, Tianchang Shen, Jiahui Huang, Huan Ling, Yifan Lu, Merlin Nimier-David, Thomas M{\"u}ller, Alexander Keller, Sanja Fidler, and Jun Gao.
\newblock {GEN3C}: 3d-informed world-consistent video generation with precise camera control.
\newblock \emph{arXiv preprint arXiv:2503.03751}, 2025.

\bibitem[{SAM 3D Team} et~al.(2025){SAM 3D Team}, Chen, Chu, Gleize, Liang, Sax, Tang, Wang, Guo, Hardin, Li, Lin, Liu, Ma, Sagar, Song, Wang, Yang, Zhang, Doll{\'a}r, Gkioxari, Feiszli, and Malik]{sam3d2025}
{SAM 3D Team}, Xingyu Chen, Fu-Jen Chu, Pierre Gleize, Kevin~J. Liang, Alexander Sax, Hao Tang, Weiyao Wang, Michelle Guo, Thibaut Hardin, Xiang Li, Aohan Lin, Jiawei Liu, Ziqi Ma, Anushka Sagar, Bowen Song, Xiaodong Wang, Jianing Yang, Bowen Zhang, Piotr Doll{\'a}r, Georgia Gkioxari, Matt Feiszli, and Jitendra Malik.
\newblock {SAM 3D}: 3dfy anything in images.
\newblock \emph{arXiv preprint arXiv:2511.16624}, 2025.

\bibitem[Shen et~al.(2024)Shen, Pi, Xia, Cen, Peng, Hu, Bao, Hu, and Zhou]{shen2024gvhmr}
Zehong Shen, Huaijin Pi, Yan Xia, Zhi Cen, Sida Peng, Zechen Hu, Hujun Bao, Ruizhen Hu, and Xiaowei Zhou.
\newblock World-grounded human motion recovery via gravity-view coordinates.
\newblock In \emph{SIGGRAPH Asia 2024 Conference Papers}, 2024.

\bibitem[{Team Wan} et~al.(2025){Team Wan}, Wang, Ai, et~al.]{wan2025}
{Team Wan}, Ang Wang, Baole Ai, et~al.
\newblock Wan: Open and advanced large-scale video generative models.
\newblock \emph{arXiv preprint arXiv:2503.20314}, 2025.

\bibitem[Tong et~al.(2026)Tong, Huang, Han, Long, and Zhu]{tong2026igrecon4ddynamicobjectawarefeedforwardreconstruction}
Yifei Tong, Ziyang Huang, Yizhao Han, Xiaoxiao Long, and Hao Zhu.
\newblock Igrecon4d: Dynamic-object-aware feed-forward reconstruction of driving scenes, 2026.

\bibitem[{Ultralytics}(2023)]{ultralytics2023yolov8}
{Ultralytics}.
\newblock {Ultralytics YOLOv8}.
\newblock Official model documentation, 2023.
\newblock Released January 10, 2023; accessed November 17, 2026.

\bibitem[Unterthiner et~al.(2018)Unterthiner, van Steenkiste, Kurach, Marinier, Michalski, and Gelly]{unterthiner2018fvd}
Thomas Unterthiner, Sjoerd van Steenkiste, Karol Kurach, Raphael Marinier, Marcin Michalski, and Sylvain Gelly.
\newblock Towards accurate generative models of video: A new metric \& challenges.
\newblock \emph{arXiv preprint arXiv:1812.01717}, 2018.

\bibitem[Van~Hoorick et~al.(2026)Van~Hoorick, Chen, Iwase, Tokmakov, Irshad, Vasiljevic, Gupta, Cheng, Zakharov, and Guizilini]{vanhoorick2026anyview}
Basile Van~Hoorick, Dian Chen, Shun Iwase, Pavel Tokmakov, Muhammad~Zubair Irshad, Igor Vasiljevic, Swati Gupta, Fangzhou Cheng, Sergey Zakharov, and Vitor~Campagnolo Guizilini.
\newblock Anyview: Synthesizing any novel view in dynamic scenes.
\newblock In \emph{European Conference on Computer Vision (ECCV)}, 2026.

\bibitem[Wang et~al.(2026)Wang, Pittaluga, Tasneem, You, Chandraker, and Jiang]{wang2026horizonforge}
Yifan Wang, Francesco Pittaluga, Zaid Tasneem, Chenyu You, Manmohan Chandraker, and Ziyu Jiang.
\newblock {HorizonForge}: Driving scene editing with any trajectories and any vehicles.
\newblock In \emph{Proceedings of the IEEE/CVF Conference on Computer Vision and Pattern Recognition}, pages 24895--24905, 2026.

\bibitem[Wang et~al.(2004)Wang, Bovik, Sheikh, and Simoncelli]{wang2004ssim}
Zhou Wang, Alan~C. Bovik, Hamid~R. Sheikh, and Eero~P. Simoncelli.
\newblock Image quality assessment: From error visibility to structural similarity.
\newblock \emph{IEEE Transactions on Image Processing}, 13\penalty0 (4):\penalty0 600--612, 2004.

\bibitem[Wen et~al.(2023)Wen, Zhao, Liu, Jia, Wang, Luo, Zhang, Wang, Sun, and Zhang]{wen2023panacea}
Yuqing Wen, Yucheng Zhao, Yingfei Liu, Fan Jia, Yanhui Wang, Chong Luo, Chi Zhang, Tiancai Wang, Xiaoyan Sun, and Xiangyu Zhang.
\newblock {Panacea}: Panoramic and controllable video generation for autonomous driving.
\newblock \emph{arXiv preprint arXiv:2311.16813}, 2023.

\bibitem[Xiang et~al.(2025)Xiang, Chen, Xu, Wang, Lv, Deng, Zhu, Dong, Zhao, Yuan, and Yang]{xiang2025trellis2}
Jianfeng Xiang, Xiaoxue Chen, Sicheng Xu, Ruicheng Wang, Zelong Lv, Yu Deng, Hongyuan Zhu, Yue Dong, Hao Zhao, Nicholas~Jing Yuan, and Jiaolong Yang.
\newblock Native and compact structured latents for 3d generation.
\newblock \emph{arXiv preprint arXiv:2512.14692}, 2025.

\bibitem[Yan et~al.(2024)Yan, Lin, Zhou, Wang, Sun, Zhan, Lang, Zhou, and Peng]{yan2024street}
Yunzhi Yan, Haotong Lin, Chenxu Zhou, Weijie Wang, Haiyang Sun, Kun Zhan, Xianpeng Lang, Xiaowei Zhou, and Sida Peng.
\newblock Street gaussians: Modeling dynamic urban scenes with gaussian splatting.
\newblock \emph{arXiv preprint arXiv:2401.01339}, 2024.

\bibitem[Yan et~al.(2025)Yan, Xu, Lin, Jin, Guo, Wang, Zhan, Lang, Bao, Zhou, and Peng]{yan2025streetcrafter}
Yunzhi Yan, Zhen Xu, Haotong Lin, Haian Jin, Haoyu Guo, Yida Wang, Kun Zhan, Xianpeng Lang, Hujun Bao, Xiaowei Zhou, and Sida Peng.
\newblock Streetcrafter: Street view synthesis with controllable video diffusion models.
\newblock In \emph{Proceedings of the IEEE/CVF Conference on Computer Vision and Pattern Recognition}, 2025.

\bibitem[Yang et~al.(2023)Yang, Ivanovic, Litany, Weng, Kim, Li, Che, Xu, Fidler, Pavone, and Wang]{yang2023emernerf}
Jiawei Yang, Boris Ivanovic, Or Litany, Xinshuo Weng, Seung~Wook Kim, Boyi Li, Tong Che, Danfei Xu, Sanja Fidler, Marco Pavone, and Yue Wang.
\newblock Emernerf: Emergent spatial-temporal scene decomposition via self-supervision.
\newblock \emph{arXiv preprint arXiv:2311.02077}, 2023.

\bibitem[Yang et~al.(2026)Yang, Fan, Shi, Peng, Wang, and Zhang]{yang2026neoverse}
Yuxue Yang, Lue Fan, Ziqi Shi, Junran Peng, Feng Wang, and Zhaoxiang Zhang.
\newblock Neoverse: Enhancing 4d world model with in-the-wild monocular videos.
\newblock \emph{arXiv preprint arXiv:2601.00393}, 2026.

\bibitem[Yu et~al.(2026)Yu, Lin, Wang, Li, Wang, Zhang, Wang, Zhou, Hu, and Peng]{yu2026infinidepth}
Hao Yu, Haotong Lin, Jiawei Wang, Jiaxin Li, Yida Wang, Xueyang Zhang, Yue Wang, Xiaowei Zhou, Ruizhen Hu, and Sida Peng.
\newblock {InfiniDepth}: Arbitrary-resolution and fine-grained depth estimation with neural implicit fields.
\newblock In \emph{Proceedings of the IEEE/CVF Conference on Computer Vision and Pattern Recognition}, pages 26920--26930, 2026.

\bibitem[Yu et~al.(2025)Yu, Hu, Xing, and Shan]{yu2025trajectorycrafter}
Mark Yu, Wenbo Hu, Jinbo Xing, and Ying Shan.
\newblock {TrajectoryCrafter}: Redirecting camera trajectory for monocular videos via diffusion models.
\newblock \emph{arXiv preprint arXiv:2503.05638}, 2025.

\bibitem[Zeng et~al.(2025)Zeng, Wu, Xiong, Wei, Guo, Zhu, Ho, Zhou, Zeng, Lu, Sun, Wang, Chen, Ye, and Zhang]{zeng2025dream4drive}
Kai Zeng, Zhanqian Wu, Kaixin Xiong, Xiaobao Wei, Xiangyu Guo, Zhenxin Zhu, Kalok Ho, Lijun Zhou, Bohan Zeng, Ming Lu, Haiyang Sun, Bing Wang, Guang Chen, Hangjun Ye, and Wentao Zhang.
\newblock Rethinking driving world model as synthetic data generator for perception tasks.
\newblock \emph{arXiv preprint arXiv:2510.19195}, 2025.

\bibitem[Zhang et~al.(2018)Zhang, Isola, Efros, Shechtman, and Wang]{zhang2018lpips}
Richard Zhang, Phillip Isola, Alexei~A. Efros, Eli Shechtman, and Oliver Wang.
\newblock The unreasonable effectiveness of deep features as a perceptual metric.
\newblock \emph{arXiv preprint arXiv:1801.03924}, 2018.

\bibitem[Zhao et~al.(2024)Zhao, Wang, Zhu, Chen, Huang, Bao, and Wang]{zhao2024drivedreamer2}
Guosheng Zhao, Xiaofeng Wang, Zheng Zhu, Xinze Chen, Guan Huang, Xiaoyi Bao, and Xingang Wang.
\newblock {DriveDreamer-2}: Llm-enhanced world models for diverse driving video generation.
\newblock \emph{arXiv preprint arXiv:2403.06845}, 2024.

\bibitem[Zhou et~al.(2023)Zhou, Lin, Shan, Wang, Sun, and Yang]{zhou2023drivinggaussian}
Xiaoyu Zhou, Zhiwei Lin, Xiaojun Shan, Yongtao Wang, Deqing Sun, and Ming-Hsuan Yang.
\newblock {DrivingGaussian}: Composite gaussian splatting for surrounding dynamic autonomous driving scenes.
\newblock \emph{arXiv preprint arXiv:2312.07920}, 2023.

\bibitem[Zhu et~al.(2025)Zhu, Zou, Jiang, Sun, Casser, Huang, Wang, Yang, Gao, Guibas, Tan, and Anguelov]{zhu2025scenecrafter}
Zehao Zhu, Yuliang Zou, Chiyu~Max Jiang, Bo Sun, Vincent Casser, Xiukun Huang, Jiahao Wang, Zhenpei Yang, Ruiqi Gao, Leonidas Guibas, Mingxing Tan, and Dragomir Anguelov.
\newblock {SceneCrafter}: Controllable multi-view driving scene editing.
\newblock \emph{arXiv preprint arXiv:2506.19488}, 2025.

\end{thebibliography}
\end{document}